\documentclass[10pt,journal,final]{IEEEtran}

\usepackage{cite}

\usepackage[comma,numbers,square,sort&compress]{natbib}
\usepackage{amsmath}
\usepackage{amssymb}
\usepackage{float}
\usepackage{amsthm}
\usepackage{graphicx}
\usepackage{epstopdf}
\usepackage{color}
\usepackage{soul}
\usepackage{verbatim}
\usepackage{tikz,times}
\usepackage{algorithmic}
\usepackage{color}
\usepackage[ruled,vlined]{algorithm2e}
  \usepackage{subfigure}
  \usepackage{footnote}
  \usepackage{multirow}

\usepackage{stackengine}
\usepackage{romannum}

\title{\LARGE \bf
	A Systematic Survey of Control Techniques and Applications in Connected and Automated Vehicles
}

\author{Wei Liu$^{\dagger}$, Min Hua$^{\dagger}$, Zhiyun Deng, Zonglin Meng, Yanjun Huang, Chuan Hu, Shunhui Song, Letian Gao, Changsheng Liu, Bin Shuai, Amir Khajepour, Lu Xiong$^*$, Xin Xia$^*$ 

\thanks{Wei Liu, Yanjun Huang, Shunhui Song,  Letian Gao, Lu Xiong, and Xin Xia are with the School of Automotive Studies, Tongji University, Shanghai, 218074, P.R. China.}
\thanks{Min Hua is with State Key Laboratory of Automotive Simulation and Control, Jilin University, Changchun, 130025, P.R. China}
\thanks{Zhiyun Deng is with the School of Mechanical Science and
Engineering, Huazhong University of Science and Technology, Wuhan, 430074, P.R. China.}
\thanks{Zonglin Meng is with the Department
of Civil and Environmental Engineering, University of California, Los Angeles,
CA 90095 USA.}
\thanks{Chuan Hu is with the School of Mechanical Engineering, Shanghai Jiao Tong
University, Shanghai, 200240, P.R. China.}
\thanks{Changsheng Liu is with 
the College of Computer Science and Technology, Zhejiang University, Hangzhou, 310027, P.R. China.}
\thanks{Bin Shuai is with the School of Vehicle and Mobility, Tsinghua University, Beijing, 100084, P.R. China.}
\thanks{Amir Khajepour is with the Department of Mechanical and Mechatronics Engineering, University of Waterloo, Waterloo, ON, Canada.}

\thanks{$\dagger$ Equal contribution. * Corresponding author: Lu Xiong, e-mail: xiong$\_$lu@tongji.edu.cn; Xin Xia, e-mail: xiaxin2000@gmail.com.}
}

\begin{document}
	\maketitle
	\thispagestyle{empty}
	\pagestyle{empty}
\begin{abstract}
Vehicle control is one of the most critical challenges in autonomous vehicles (AVs) and connected and automated vehicles (CAVs), and it is paramount in vehicle safety, passenger comfort, transportation efficiency, and energy saving. This survey attempts to provide a comprehensive and thorough overview of the current state of vehicle control technology, focusing on the evolution from vehicle state estimation and trajectory tracking control in AVs at the microscopic level to collaborative control in CAVs at the macroscopic level. First, this review starts with vehicle key state estimation, specifically vehicle sideslip angle, which is the most pivotal state for vehicle trajectory control, to discuss representative approaches. Then, we present symbolic vehicle trajectory tracking control approaches for AVs. On top of that, we further review the collaborative control frameworks for CAVs and corresponding applications. Finally, this survey concludes with a discussion of future research directions and the challenges. This survey aims to provide a contextualized and in-depth look at state of the art in vehicle control for AVs and CAVs, identifying critical areas of focus and pointing out the potential areas for further exploration.

\textit{Index Terms-}Autonomous vehicles; Connected and automated vehicles; State estimation; Vehicle sideslip angle; Trajectory tracking control; Collaborative control

\end{abstract}


\section{Introduction}

\IEEEPARstart{A}{utomated} driving and collaborative driving automation technologies are revolutionizing future transportation systems regarding
reducing traffic congestion, enhancing safety, and improving energy efficiency \cite{xu2021opencda, kang2019test, xu2023v2v4real, paden2016survey, huang2022survey, ju2022survey, tian2022federated, ma2020statistical, luo2022multisource, hu2022review, miao2021data, hu2020fuzzy, ma2022verification}. AVs and CAVs as the key instantiation of automated driving and collaborative driving automation become more and more widely deployed and they benefit both individuals and society \cite{hancock2019future, su2022uncertainty}. 
In the past few decades, manufacturers and scholars have been thriving in automotive industrialization to maximize benefits from automated driving and collaborative driving automation \cite{bhavsar2017risk, he2023data, he2020data, xu2021holistic, shi2023even, zhang2022hierarchical, xin2017vehicle}. The hierarchical framework, including perception, planning, and control modules, is the popular pipeline for both automated driving and collaborative driving automation techniques. One of the most challenging tasks for automated driving and collaborative driving automation is to develop safe and efficient vehicle control modules for both AV and CAV \cite{gao2021autonomous}. To clarify, vehicle control is to make an AV or CAV follow a desired route or trajectory \cite{katrakazas2015real}. More specially, for AVs, we will focus on vehicle trajectory tracking control at the microscopic level, while more efforts on collaborative control will be put for CAVs at the microscopic level \cite{kiran2021deep}.

\begin{figure*}[ht]
\centering
\includegraphics[width=1\linewidth]{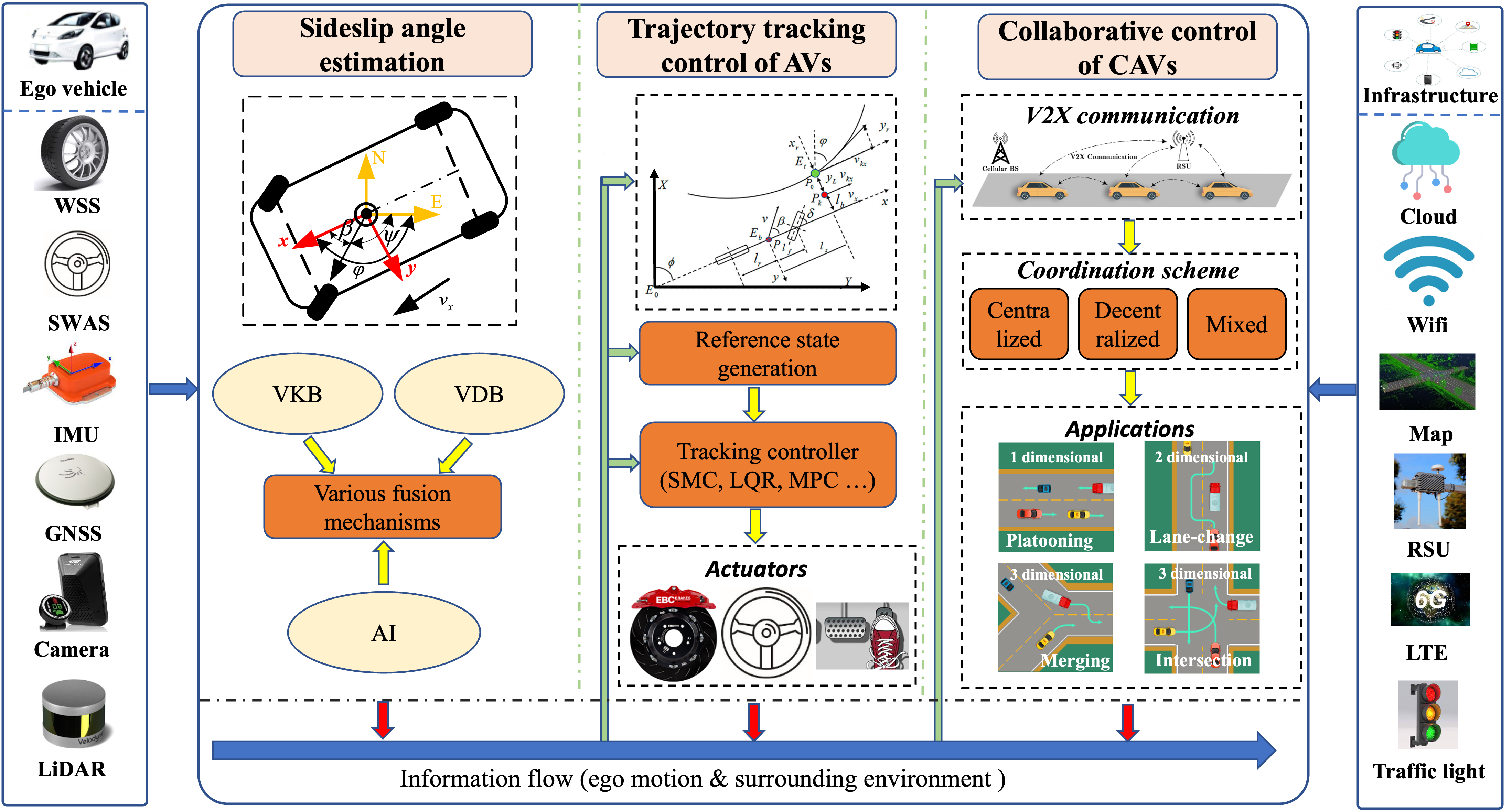}
\caption{Overall framework of vehicle control for AVs and CAVs. AV, autonomous vehicle; CAV, connected and automated vehicle; WSS, wheel speed sensor; SWAS, steering wheel angle sensor; IMU, inertial measurement unit; GNSS, global navigation satellite system; VKB, vehicle-kinematics-based; VDB, vehicle-dynamics-based; AI, artificial intelligence; SMC, sliding model control; LQR, linear quadratic regulation; MPC, model predictive control; RSU, roadside unit; LTE, long-term evolution}
\label{fig: overall_framework}
\end{figure*}

Before diving into vehicle control, it is worth mentioning that accurate and reliable vehicle state information is the prerequisite and essential for vehicle control in AVs and CAVs \cite{xia2022estimation, xia2016estimation}. However, key states such as longitudinal and lateral velocity, sideslip angle, orientation, and tire forces cannot be directly measured for commercial vehicles. These vital states can only be estimated indirectly. Among these states, the sideslip angle is one of the most important states of interest as it comprises both longitudinal and lateral velocity information, i.e., the sideslip angle accuracy is an indicator of the accuracy of the longitudinal velocity and lateral velocity estimation accuracy. Despite the importance of the sideslip angle, it is also the most comprehensive but difficult to be estimated among the states mentioned above because errors from other states such as longitudinal velocity, and roll angle contribute to the error in sideslip angle \cite{xiong2020imu}. In addition, the sideslip angle is critical for vehicle stabilization, motion planning, road condition estimation, handover modules, and vehicle navigation \cite{xia2022autonomous}. Namely, without an accurate sideslip angle, the performance of the applications mentioned above will be compromised and even sustain failure inevitably. Thus, efficient and robust vehicle sideslip angle estimation is of much importance and has been attracting much attention during the past few years in the vehicle control community. Although other states such as longitudinal speed information are also necessary for vehicle control applications, compared to sideslip angle, it is easier to obtain and much literature has well-addressed its estimation \cite{zhuoping2009review}. Therefore, in this work, we focus on the sideslip angle estimation and with the significance of sideslip angle to vehicle control, it is necessary to tackle this sideslip angle estimation problem by providing the readers with a review of this aspect such that vehicle control can be discussed thoroughly. Based on the literature, using the accessible information from cheap onboard sensors, including the steering wheel angle sensor (SWAS), wheel speed sensors (WSS), the inertial measurement unit (IMU), and global navigation satellite systems (GNSS), sideslip angle estimation can be categorized into three main approaches: onboard-sensor-based (OSB) approach, GNSS-augmented (GAU) approach, and artificial-intelligence-augmented (AIA) approach \cite{liu2021automated, xia2021autonomous}. Each of the approaches will be elaborated on in the next section.

Founded on state estimation, in particular side slip angle estimation, for an individual AV at the microscopic level, the vehicle trajectory tracking control is designed to track the desired paths under diverse driving scenarios. Numerous control strategies, including PID, linear quadratic regulator (LQR), feed-forward and feedback control, robust control, sliding mode control (SMC), model predictive control (MPC), and learning-based method \cite{shan2015cf, ding2017sliding, lu2023gain, zhou2019multiobjective} have been developed to adapt to the challenge scenarios such as driving on low friction road with excessive steering. Despite the substantial progress of model-based control algorithms based on PID, LQR, SMC, MPC, etc., there still remain corner cases where it is difficult to control the vehicle. Two potential issues from the vehicle dynamic models compromise the vehicle control algorithms normally: 1) tire cornering characteristics are subject to strong nonlinearities and uncertainties, particularly in extreme driving conditions; 2) the model accuracy and computation workload should be a trade-off in the real application. To fill this gap and make the control algorithm more robust, recently, learning-based methods have gained attention due to the numerous appealing results that have been achieved in many fields, such as intelligent systems control, decision-making, and prediction with the characteristics of the desirable self-optimization and adaptability 
\cite{chen2022explain, chen2022relax, chen2021costly, herobust, han2022solution, he2022robust, dou2022sampling, han2022stable}. Although learning-based methods are well suitable to control problems in some complex and dynamic environments, they lack model interpretability. As a consequence, it is challenging for learning-based methods to guarantee feasibility in the real world in the current stage. There is a need for such a review work discussing both the model-based and model-free vehicle trajectory tracking control methods.

The limitations of individual AVs to continuously sense the dynamic and uncertain environment and without planning the behavior of AVs at a macroscopic level have led to growing interest in CAVs development within the realm of collaborative driving automation in intelligent transportation systems (ITS). Leveraging the shared relevant information and planned trajectories with surrounding AVs' intention beyond line-of-sight and field-of-view will further boost the robustness, safety and efficiency of CAVs, leading to safer and more efficient transportation \cite{hancock2019future,zeng2019joint}. Especially, collaborative control, one of the key features of collaborative driving automation, is an important component of CAV's development and plans the behavior of individual CAV. 
Previous studies on collaborative control can be classified into three main control approaches: centralized, decentralized, and mixed control approaches, each with its advantages and limitations for achieving efficient and safe operation of CAVs. These methods differ in terms of the level of control authority, information sharing, and decision-making processes among CAVs, ranging from a central authority controlling all vehicles in the centralized mode to each vehicle autonomously making its own decisions based on local information in the decentralized mode, and a combination of both in the mixed control mode.
To ensure practical applicability, most existing research focuses on specific driving scenarios and traffic facilities, such as platooning, lane change, merging, and intersection management. Thus, existing reviews or surveys mainly discuss control techniques of  CAVs in various scenarios. In \cite{montanaro2019towards}, five use cases from CAVs: vehicle platooning, lane change, intersection management, energy management, and road friction estimation, have been investigated to achieve potential benefits in the current road transportation system. In \cite{aradi2020survey}, a comprehensive review on five selected subjects has been conducted for CAVs: inter-CAV communications; security of CAVs; intersection control for CAVs; collision-free navigation of CAVs; and pedestrian detection and protection. In \cite{guanetti2018control}, this survey presents a control and planning architecture for CAVs and analyses the state of the art on each functional block. It mainly focuses on energy efficiency strategies. Sarker et al. thoroughly investigate three fundamental and interconnected areas of CAVs: sensing and communication technologies, human factors, and information-aware controller design \cite{sarker2019review}. To this end, they suffer from the following two drawbacks: 1) the importance of state estimation for robust vehicle control is overlooked; 2) The vehicle control of AVs and CAVs has been discussed separately rather than in a cohesive manner. Aiming to address these two issues, this survey covers several aspects of vehicle control techniques and applications evolving from AVs to CAVs. 

The overall framework for this survey is shown in Fig.~\ref{fig: overall_framework}. With the multi-modal sensors including on-board sensors, IMU, GNSS, camera, and LiDAR as inputs on AVs or CAVs, state estimation techniques especially sideslip angle considering diverse sensors configuration and model features, will be reviewed in section \uppercase\expandafter{\romannumeral2}. Then, classical vehicle physical models and trajectory tracking control algorithms of AVs will be discussed in Section \uppercase\expandafter{\romannumeral3}. In addition, for collaborative control of CAVs, the enabling techniques, critical components, methodologies, and potential applications are surveyed in Section \uppercase\expandafter{\romannumeral4}. Section \uppercase\expandafter{\romannumeral5} concludes with remarks on the state and potential areas for future research.

\section{Vehicle Sideslip Angle Estimation}
\begin{figure}[ht]
\centering
\includegraphics[width=1\linewidth]{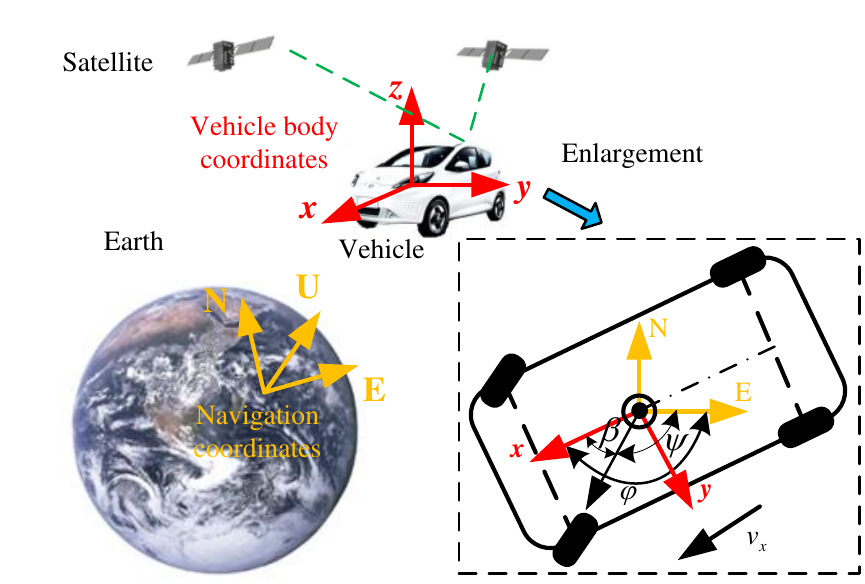}
\caption{For coordinates, E, N, and U denote the east, north, and upward directions, respectively. $x$, $y$, and $z$ denote the vehicle body's forward, left, and upward directions. $\beta$ is the sideslip angle of the vehicle which is the angle between its velocity direction and the heading of the vehicle, $\psi$ is the course of the vehicle which is the vehicle velocity direction, and $\phi$, $\theta$, and $\varphi$ are the roll, pitch, and heading angle of the vehicle concerning the east direction in the navigation coordinate.}
\label{fig:Coordinates}
\end{figure}

Vehicle state information plays a critical role in the motion planning, decision-making, and control techniques of AVs and CAVs \cite{liu2018vehicle, liu2017vehicle}. The states comprise vehicle position, velocity (longitudinal and lateral velocity in the vehicle coordinate), and attitudes (vehicle roll, pitch, and yaw). Fig.~\ref{fig:Coordinates} illustrates the definitions of the aforementioned states, where the superscripts $n$ and $b$ represent the navigation and body coordinates, respectively. In the navigation frame, $x$, $y$, and $z$ point east, north, and upward, and in the body frame, they point forward, left, and upward. As shown in (\ref{sideslip}), accurate knowledge of the longitudinal and lateral velocities is required to calculate the vehicle sideslip angle, which is crucial for vehicle control. While expensive equipment, such as the RT3000 and Kistler S-Motion, can measure some of these state variables, they are not feasible for mass-produced vehicles. Therefore, many scholars have been working on estimating the vehicle sideslip angle for practical automobile applications. It should be noted that although the localization/positioning estimation of the AVs and CAVs belongs to the scope of state estimation for AVs and CAVs, it is more related to the navigation/surveying and mapping community and extensive and comprehensive reviews have been conducted in \cite{li2020toward, jing2022integrity, zhang2020required, chen2021ginav}. To make this review work dedicatedly serve the interest of the vehicle control community, we focus on the aspect of vehicle state estimation, i.e., sideslip angle estimation. In another aspect, the sideslip angle estimation is equivalent to the heading angle estimation, which is one of the most critical states in the navigation system \cite{farrell1999global}. The relationship between the sideslip angle and heading angle has been revealed in \cite{xia2021autonomous} as follows: 

\begin{equation}
\beta = tan^{-1}{(\frac{v_y}{v_x})}
\label{sideslip}
\end{equation}

Currently, the design of vehicle state estimation is dominated by Kalman filters (KFs) and nonlinear observers. KFs are widely used due to their simplicity and robustness. They consist of two phases: the prediction phase and the correction phase. The prediction phase produces estimates of the current state variables and their uncertainties. When the measurement signal arrives, the estimated variables are corrected in the correction phase. The prediction phase can be described by equation (\ref{prediction stage}).

\begin{figure*}[ht]
\centering
\includegraphics[width=1\linewidth]{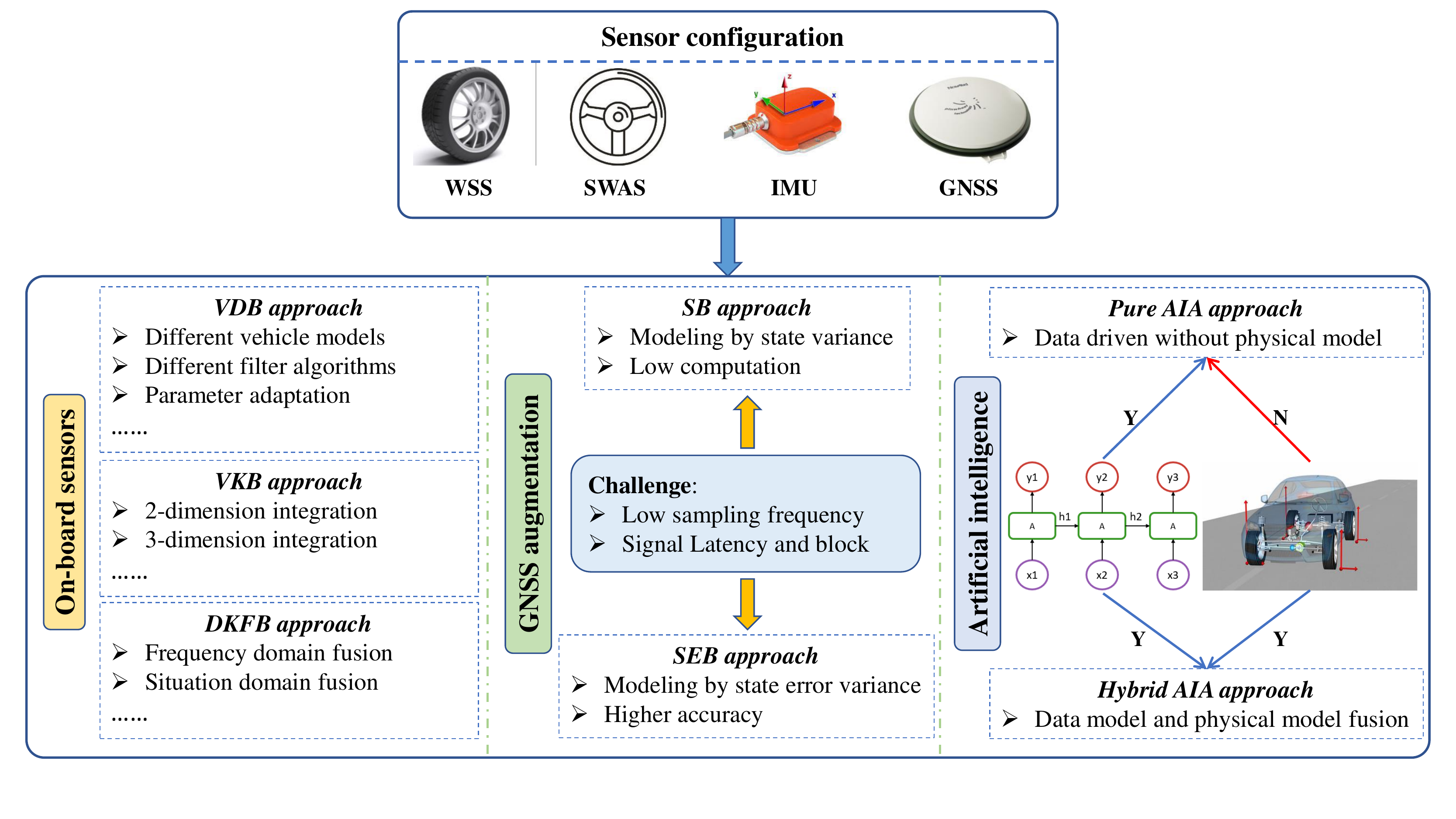}
\caption{Summary vehicle sideslip angle estimation approaches. WSS, wheel speed sensor; SWAS, steering wheel angle sensor; IMU, inertial measurement unit; GNSS, global navigation satellite system; VKB, vehicle-kinematics-based; VDB, vehicle-dynamics-based;  DKFB, dynamic-kinematics-fusion-based; SB, state-based; SEB, state error-based; AIA, artificial-intelligence-augmented.}
\label{fig: State_classification}
\end{figure*}

\begin{equation}
    \begin{split}
    &x_{k+1 \mid k}=\boldsymbol{\Phi}_k x_{k \mid k}+H\cdot u_k \\
    &\boldsymbol{P}_{k+1 \mid k}=\boldsymbol{\Phi}_k \boldsymbol{P}_{k \mid k} \boldsymbol{\Phi}_k^{\mathrm{T}}+Q_k
    \end{split}
    \label{prediction stage}
\end{equation}
where $\boldsymbol{P}$ is the error covariance matrix; $x_{k+1 \mid k+1}$ is the output estimated value; $\boldsymbol{\Phi}_k$ is the state transition matrix; $Q_k$ is the covariance of the process noise.

The correction phase is described as follows:
\begin{equation}
    \begin{split}
    &K_{k+1}=\boldsymbol{P}_{k+1 \mid k} \boldsymbol{C}^{\mathrm{T}}\left(\boldsymbol{C} \boldsymbol{P}_{k+1 \mid k} \boldsymbol{C}^{\mathrm{T}}+R_{k+1}\right)^{-1} \\
    &x_{k+1 \mid k+1}=x_{k+1 \mid k}+K_{k+1}\left(y_{k+1}-\boldsymbol{C} x_{k+1 \mid k}\right) \\
    &\boldsymbol{P}_{k+1 \mid k+1}=\left(I-K_{k+1} \boldsymbol{C}\right) \boldsymbol{P}_{k+1 \mid k}
    \end{split}
    \label{correction stage}
\end{equation}
where $K$ is the feedback gain of the KF; $x_{k+1 \mid k+1}$ is the output estimated value; $\boldsymbol{C}$ is the measurement transition matrix; $R_{k+1}$ is the covariance of the measurement noise.

However, using the standard KF for state estimation under high dynamic conditions is suboptimal. To overcome this limitation, some researchers have proposed variants of the KF to improve its performance. Alternatively, others have developed nonlinear observers to estimate vehicle states, taking into account the strong nonlinearity of the vehicle dynamic model during extreme driving conditions. Unlike KFs, there is no fixed design paradigm for nonlinear observers. In particular, feedback coefficients are calculated by constructing a subtle Lyapunov function and restricting the function derivatives to be less than zero \cite{gao2019multi}.

Fig.~\ref{fig: State_classification} presents the detailed classification of the vehicle sideslip angle estimation. In general, it can be categorized into three main approaches: OSB approach, GAU approach, and AIA approach. The OSB approach can be further classified into three kinds: vehicle-kinematics-based (VKB) approach, vehicle-dynamic-based (VDB) approach, and dynamic-kinematics-fusion-based (DKFB) approach. The GAU approach can be further classified into the state-based (SB) approach and the state-error-based (SEB) approach. The AIA approach can also be classified into the hybrid AIA approach and the pure AIA approach.

\subsection{OSB approach}
Onboard sensors, such as SWAS, WSS, and IMU, are essential for ensuring safe vehicle operation. Researchers have developed various mathematical models that utilize the input data from these sensors to estimate the vehicle's sideslip angle in real time. These comprehensive estimation approaches can be categorized into three main types: the VDB approach, the VKB approach, and the DKFB approach. In the following sections, we will provide a detailed overview of each of these approaches.

\subsubsection{VDB approach}
Considering that the vehicle dynamic model has a relatively low dependence on sensors' accuracy, substantial research on vehicle sideslip angle estimation based on vehicle dynamic models has been conducted in industry and academia in the past two decades. The performance of the VDB approach is coupled with the reliability of the vehicle dynamic model to a great extent. Accordingly, the degree of freedom (DOF), the road friction, the nonlinear characteristics of the tire, and the uncertainty of the vehicle model parameters will impact the accuracy of the estimation algorithms. To facilitate an overview of the VDB approach, a classical vehicle dynamic model is presented here as an example to describe the vehicle motion state. A 2-DOF vehicle model can be written as follows:
\begin{equation}
    \begin{split}
    &\dot{\beta}=\frac{F_{y f} \cdot \cos \delta+F_{y r}}{m v_x}-\dot{\varphi} \\
    &\ddot{\varphi}=\frac{l_f \cdot F_{y f} \cdot \cos \delta-l_r F_{y r}}{I_z}
    \end{split}
    \label{2-DOF}
\end{equation}
where $m$ is the vehicle mass; $\dot{\beta}$ and $\ddot{\varphi}$ are the derivative of the sideslip angle and yaw rate, respectively; $F_{y f}$ and $F_{y r}$ are the lateral force of the front and rear tires; $\delta$ is the steering wheel angle; $m$ is the vehicle mass; $I_z$ is the vehicle yaw moment of inertia; $l_f$ and $l_r$ is the distance from the center of gravity (COG) to the front axle and rear axle; $v_x$ is the vehicle longitudinal velocity.

Based on the classical vehicle dynamic model and its variants, enormous VDB approaches and a great variety of estimators have been proposed. 
Among them, the extended Kalman filter (EKF) method has received significant attention, as evidenced by a number of studies \cite{dakhlallah2008tire, hrgetic2014vehicle, li2014variable, sun2022research, jin2015estimation, nam2012lateral, jiang2017novel}. For instance, Dakhlallah et al. propose a sideslip angle estimator based on the Dugoff tire forces model, using the EKF approach \cite{dakhlallah2008tire}. Hrgetic et al. introduce stochastic modeling of tire forces to design an EKF-based sideslip angle estimator \cite{hrgetic2014vehicle}. Meanwhile, Li et al. use the sideslip angle rate as the feedback measurement and design an EKF based on steering torque to achieve a speedy response in estimating the sideslip angle \cite{li2014variable}. However, the assumption of Gaussian noise in EKF may introduce extra estimation errors. Thus, Sun et al. introduce a noise update module to adaptively update the noise \cite{sun2022research}. Jin et al. construct linear and nonlinear tire models and employ an interacting multiple-model filter with EKF to estimate the sideslip angle seamlessly \cite{jin2015estimation}. Nam et al. introduce lateral tire force sensors as the augmented measurement signal to estimate the vehicle sideslip angle \cite{nam2012lateral}. Additionally, Jiang et al. propose a 4-DOF model with roll motion to estimate the vehicle sideslip angle, given that the roll angle can corrupt the vehicle's lateral acceleration measurement \cite{jiang2017novel}.


When vehicle tires reach the extreme nonlinear region, the first-order linearization used in EKF may not be accurate enough for state estimation. In contrast, unscented Kalman filtering (UKF) can work directly with nonlinear models and estimate states using a set of sigma points to avoid local linearization, making it a suitable approach for vehicle sideslip angle estimation. This is demonstrated in several studies, including \cite{doumiati2010onboard, doumiati2009unscented, wielitzka2014state, strano2018constrained, jin2020online, bertipaglia2022two}. For example, \cite{doumiati2010onboard, doumiati2009unscented, wielitzka2014state} uniformly demonstrate that UKF outperforms EKF in estimating vehicle sideslip angle estimation. Due to the boundaries on state variables, Strano et al. introduce the constrained UKF to improve convergence and estimation error performance \cite{strano2018constrained}. Jin et al. deduce the local observability of UKF through differential geometry theory \cite{jin2020online}. Bertipaglia et al. address the challenge of calibrating the process noise matrix of UKF by introducing a two-stage Bayesian optimization method \cite{bertipaglia2022two}. In addition to EKF and UKF, particle filter (PF) and cubature Kalman filter (CKF) have also been employed for precise estimation of the sideslip angle. PFs, like UKFs, do not rely on local linearization techniques or rough approximations. Chu et al. estimate vehicle sideslip angle through an unscented PF \cite{chu2015wheel}. Wang et al. propose robust CKF for precise estimation of the sideslip angle \cite{wang2020estimation}.
At the same time, research on vehicle sideslip angle estimation also includes nonlinear observers apart from KFs and their variants. Zhang et al. propose a gain-scheduling observer for sideslip angle estimation, accounting for the uncertainty of 2-DOF vehicle dynamic models \cite{zhang2015robust}. Gao et al. design a high-gain observer with input-output linearization for vehicle sideslip angle estimation \cite{gao2010sideslip}. To address the issue of time-varying speed, Zhang et al. present a Takagi-Sugeno (T-S) observer \cite{zhang2016novel}. In addition, You et al. develop a nonlinear observer to estimate vehicle sideslip angle and road bank angle simultaneously, taking into account that the measurement of lateral acceleration is affected by vehicle roll angle and road bank angle \cite{you2009new}.

As illustrated above, VDB approaches require vehicle parameters. However, certain parameters, such as COG and tire cornering stiffness, are highly dependent on driving and load conditions. Uncertainty in these parameters can introduce estimation errors and even cause the estimator to fail to converge.

\subsubsection{VKB approach}
Given the drawbacks of inaccurate vehicle models, especially during critical driving conditions, some scholars have opted to forego VDB approaches and instead utilize vehicle sensor information, such as IMU, WSS, and SWAS, to develop VKB approaches. A dominated 2-DOF vehicle kinematic model in the plane can be expressed as follows:
\begin{equation}
    \begin{aligned}
    & \dot{v}_x(t)=a_x(t)+\dot{\varphi} v_y \\
    & \dot{v}_y(t)=a_y(t)-\dot{\varphi} v_x
    \end{aligned}
    \label{kinematic_2D}
\end{equation}
Through direct integration \cite{best1998real}, the sideslip angle could be derived as follows: 
\begin{equation}
    \beta=\beta_{0}+\int\left(\frac{a_{y}}{v}-\dot{\psi}\right) d t
    \label{integration}
\end{equation}

Unfortunately, this method may produce integration errors, and in severe situations, it could even result in the estimate diverging. To eliminate the above issue, Kim et al. design an EKF to estimate the vehicle sideslip angle considering time-varying longitudinal and lateral velocity changes \cite{kim2011sideslip}. Nevertheless, the authors don't specify how to obtain the measured longitudinal and lateral velocity for feedback correction. To suitably process the IMU measurements and eliminate the undesired effect, Selmanaj et al. encompass vehicle longitudinal speed, the sensor offsets, the vehicle roll angle, and the accelerations of COG \cite{selmanaj2017robust, selmanaj2017vehicle}.

Based on the literature reviewed above, it is evident that VKB approaches rely on the IMU as the core sensor. Anyhow, the IMU output signal is susceptible to temperature drift, bias error, random noise, and the gravity component resulting from the roll and pitch angle. To obtain velocity accurately by integrating the sensor's acceleration, it is essential to remove bias, noise, and the gravity component. Otherwise, prolonged integration can lead to error accumulation.

\subsubsection{DKFB approach}
Since both VDB and VKB approaches have limitations, the question of how to combine them into a complementary DKFB approach remains open. According to \cite{piyabongkarn2008development}, VDB approaches exhibit a higher level of confidence at low frequencies, while VKB approaches perform better at high frequencies. Accordingly, a first-order filter can be employed to integrate the VKB and VDB approaches. The fusion result can be obtained as follows:

\begin{equation}
    \hat{\beta}=\frac{1}{\tau s+1} \hat{\beta}_{\text {model }}+\frac{\tau}{\tau s+1} \hat{\beta}_{\text {kin }}
    \label{frequency}
\end{equation}
where $\tau$ is a filter parameter; $\hat{\beta}_{\text {model }}$ is the estimation result with the VDB approach; and $\hat{\beta}_{\text {kin }} $ is the estimation result with the VKB approach. 

Chen et al. \cite{chen2008sideslip} utilized a planar kinematics model to obtain the sideslip angle, where the lateral acceleration variable is determined by the lateral dynamic equation. Notably, the observability of the model is weak when the yaw rate is small. In this case, only the prediction phase is performed. 
Building on this work, Liao et al. employ a kinematics observer to adapt the tire cornering stiffness and improve the OSB approach's performance in the nonlinear tire region \cite{liao2019adaptive}. Chen et al. design a VKB approach based on the integral method, which provides relatively accurate results in a short period \cite{chen2016ukf}. Wang et al. integrate a UKF-based 3-DOF dynamics model and an EKF kinematic model using a weighting factor to estimate the vehicle sideslip angle \cite{wang2019vehicle}. Additionally, Chen et al. fuse VDB with CKF and VKB with integration to assess the vehicle sideslip angle. The corresponding coefficient weight is determined by the vehicle's non-linearity degree and tire-road coefficient \cite{cheng2017fusion}. Finally, Li et al. develop a weight allocation strategy to fuse the VKB and VDB models, combining the front wheel steering angle, transient features of lateral acceleration, and yaw rate  \cite{li2020fusion}.

With increasingly advanced sensor technology, the acceleration and angular velocity signals acquired by the IMU are gradually being upgraded from just a two-dimensional plane to a three-dimensional space. This improvement allows for the establishment of the relationship between rotation and translation of the vehicle in three dimensions. The corresponding model can be expressed as follows:
\begin{equation}
    \begin{split}
    \left[\begin{array}{c}
    \dot{v}_{x} \\
    \dot{v}_{y} \\
    \dot{v}_{z}
    \end{array}\right] = &\left[\begin{array}{c}
    a_{x_{r}} \\
    a_{y_{r}} \\
    a_{z_{r}}
    \end{array}\right]-\left[\begin{array}{ccc}
    0 & -\dot{\varphi}_{r} & \dot{\theta}_{r} \\
    \dot{\varphi}_{r} & 0 & -\dot{\phi}_{r} \\
    -\dot{\theta}_{r} & \dot{\phi}_{r} & 0
    \end{array}\right]\left[\begin{array}{l}
    v_{x} \\
    v_{y} \\
    v_{z}
    \end{array}\right]-\\ &\left[\begin{array}{c}
    -g \sin \theta \\
    g \sin \phi \cos \theta \\
    g \cos \phi \cos \theta
    \end{array}\right]
    \end{split}
    \label{kinematic_3D}
\end{equation}
where the subscript $r$ is real value.

With the above information, the acceleration generated by gravity components with changing attitudes can be detached. Benefiting from this motivation, Xia et al. fuse three-dimensional estimators to obtain vehicle sideslip angle \cite{xia2018automated}. In minor excitation conditions, the VDB approach assists the VKB approach in estimating the sideslip angle, while in large excitation, the sideslip angle estimation would rely only on the VKB approach. Furthermore, Xiong et al. propose a novel DKFB approach considering the lever arm between the IMU and COG based on the driving conditions \cite{xiong2019imu}. The proposed architecture consists of the following four estimators: lateral and longitudinal velocity estimators with the VDB approach, and attitude/velocity estimators with the VKB approach. The driving conditions determine the fusion mechanism. More importantly, the VDB approach can be used as the feedback signal for the VKB approach when the vehicle is under normal driving conditions. However, in critical driving conditions, the VDB approach is cut off, and only the VKB approach is used to estimate the vehicle sideslip angle. The performance of the proposed method is verified by double lane change and slalom maneuvers.

\subsection{GAU approach}
With the increasing development of AVs and CAVs, there are now more sensors available to provide input information for efficient state estimation, including GNSS, cameras, and LiDAR \cite{xiang2022v2xp, xu2022cobevt, xu2022towards, swerdlow2023street, liu2020vision}. As GNSS is commonly used in AVs and CAVs with low cost, there has been extensive research on estimating the vehicle sideslip angle using GAU approaches that incorporate GNSS position and velocity information in navigation coordinates. Depending on the characteristics of the variables in the state equation, these approaches can be further divided into the following two categories: the SB approach and the SEB approach.

\subsubsection{SB approach}

The SB approach directly relies on variable states in the model. The advantage of this approach lies in its simplicity of design and derivation, as well as its relatively low computational requirements. Therefore, it is widely adopted in vehicle state estimation. Naets et al. propose an EKF that estimates vehicle sideslip angle using lateral velocity, yaw rate, longitudinal velocity, and tire cornering stiffness as the states vector \cite{naets2017design}. They also analyze four types of measurements for feedback correction, including yaw rate, lateral acceleration, four-wheel speed, and GNSS velocity. Additionally, Katriniok et al. assess the local observability of the estimation framework with the help of GNSS-based horizontal velocity \cite{katriniok2015adaptive}. Park et al. synthesize VDB and VKB approach with EKF to attain vehicle sideslip angle, and GNSS velocity measurement corrects the VKB approach \cite{park2018integrated}. Liu et al. construct the kinematic information between GNSS and IMU to estimate vehicle sideslip angle based on a nonlinear observer, in which the observability and convergence rate of the observer is verified \cite{liu2018intelligent}. Similarly, Ding et al. calculate the sideslip angle based on EKF but ignore the time synchronization problem among different sensors \cite{ding2021event}. To fill the performance gap due to time synchronization, Yoon et al. propose a kinematic-based estimator to obtain the vehicle sideslip angle by merging IMU with two GNSS receivers \cite{yoon2013cost}. Liu et al. develop an observer-predictor with multi-sensor fusion to handle the issue of measurement delay for GNSS and camera, and they adopt an adaptive fading KF to enhance the yaw and roll angle estimation performance \cite{liu2018intelligent}. Besides time synchronization, the low sampling frequency of GNSS is another critical issue. The control period for vehicle stability is typically 10 ms or 20 ms, corresponding to 100 Hz or 50 Hz, respectively. Nonetheless, the sampling frequency of GNSS is usually much less than 50 Hz, causing the accumulation of integrated errors when GNSS is unavailable. To mitigate this issue, Liu et al. propose a novel inverse smoothing and gray prediction fusion algorithm. Additionally, the authors employ a three-dimensional vehicle model with higher fidelity to decouple the gravity component of acceleration signals \cite{liu2021automated}.

\subsubsection{SEB approach}
\begin{figure}[h!]
\centering
\includegraphics[width=1\linewidth]{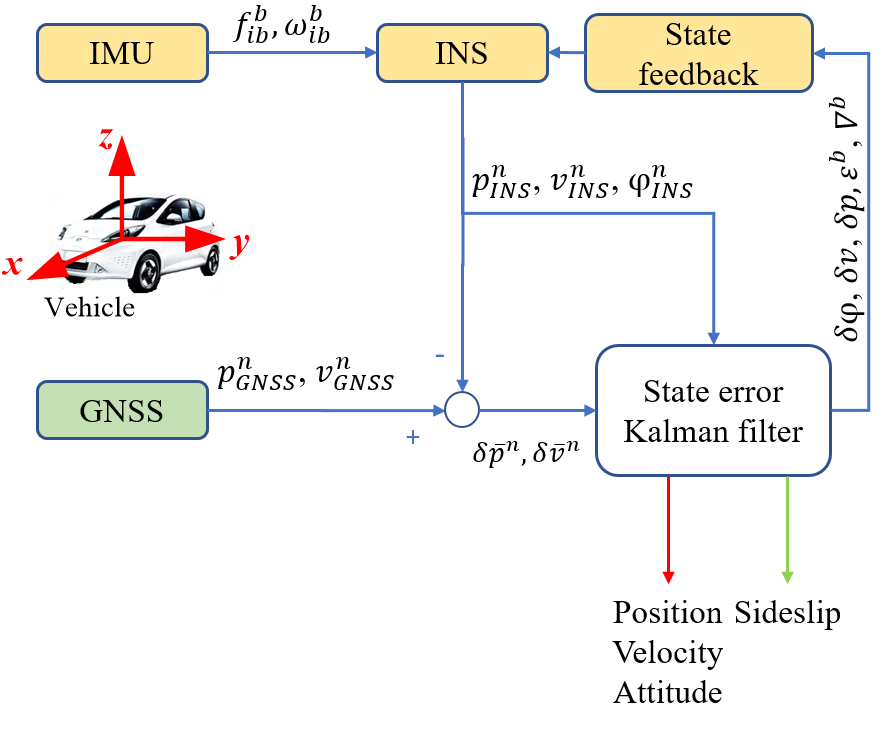}
\caption{{INS and GNSS fusion scheme. IMU, inertial measurement unit; GNSS, global navigation satellite system; INS, inertial navigation system.}}
\label{fig:INS calculation}
\end{figure}

Although the SB with GAU approaches improve the accuracy of sideslip angle estimation, they still could not accurately estimate the bias of the gyroscope and accelerometer. To alleviate the aforementioned limit, some scholars also adopt an alternative GNSS and IMU fusion strategy, called the SEB approach, which is based on the inertial navigation system (INS) framework. SEB approaches offer improved accuracy in attitude estimation and greater robustness compared to SB approaches \cite{zhuang2023multi}. SEB approaches have an advantage in that they use state errors (e.g., attitude, velocity, and position) instead of directly using the state vector. This leads to errors close to the origin, avoiding problems like gimbal locking and singular parameters. An example INS framework for sideslip angle estimation is shown in Fig.~\ref{fig:INS calculation}. The position error and velocity error derived from GNSS and INS, along with their covariance matrices, are used as the measurements for the KF. The state vector of the KF typically contains position error, velocity error, attitude error, gyro bias, and acceleration bias. Over the past ten years, various works related to vehicle sideslip angle estimation have been carried out based on the aforementioned model variants.
Xiong et al. propose a parallel innovation adaptive estimator to address the low sampling rate issue of GNSS measurements for sideslip angle estimation \cite{xiong2020imu}. Xia et al. present a hybrid strategy to address the INS heading error for sideslip angle estimation \cite{xia2021autonomous}. Considering that the heading error of INS coupled by vehicle sideslip angle is not well observable, Xia et al. introduce the VDB approach to augment the heading error into the velocity measurements \cite{xia2021advancing, xia2021vehicle}. Xia et al. apply the heading angle measurement from the dual-antenna GNSS to compensate for the course angle offset, hereby improving vehicle sideslip angle estimation \cite{xia2021vehicleMSSp}. To eliminate the error from yaw misalignment, Xia et al. construct the novel linear attitude and velocity error dynamic models and design a weighting scheme to fuse the SEB and VDB approach based on the vehicle lateral excitation level to enhance the sideslip angle estimation performance \cite{xia2022estimation}. Furthermore, Xia et al. investigate the stability of the proposed sideslip angle estimation algorithm \cite{xia2022autonomous}. Zhang et al. present an innovative VKB method for sideslip angle estimation \cite{zhang2022precise}, which continuously estimates the attitude and velocity of the vehicle to obtain the sideslip angle. Within the INS-based framework, the sideslip angle is the most difficult to estimate due to the observability issue \cite{xia2021autonomous}. To some extent, the sideslip angle estimation is equivalent to  the heading angle estimation in the GNSS/INS integration system which is dedicated to vehicle localization \cite{xia2021advancing}. In another manner, resolving the sideslip angle estimation problem through multisensor fusion considering vehicle dynamics can benefit the typical GNSS/INS integration system to provide the vehicle position in the navigation coordinate. And the position information is of great importance for collaborative control of CAVs \cite{xia2022automated, xu2023opencda, xu2022v2x}. Next, we also briefly summarize the acquisition of vehicle position information through the SEB approach. Gao et al. propose a vehicle navigation algorithm aided by vehicle sideslip angle estimation \cite{Gao2022Improve}. Similarly, Lu et al. leverage the GNSS course angle to enhance vehicle heading angle estimation, hereby improving vehicle position estimation \cite{lu2022vehicle}. Xu et al. also introduce the sideslip angle as the correction feedback to improve vehicle navigation during GNSS outages \cite{xu2018enhancing}. Several comprehensive surveys and significant recent contributions on the subject of vehicle navigation itself are discussed in \cite{jing2022integrity, zhang2020required, chen2021ginav}. 
From this point of view, the vehicle sideslip angle can improve the accuracy of vehicle navigation, which in turn has a positive impact on the collaborative control of CAVs,

\subsection{AIA approach}
Over the past decade, AI technology has made significant advances \cite{hu2020dasgil, liu2022yolov5, li2021metasaug, liu2020afnet, Song_2023_WACV, xiang2022tkil, rao2022quadformer, liu2022psdc, zheng2020bi, li2023domain, li2022learning, cong2022stcrowd}. This progress has motivated some researchers to apply AI for vehicle sideslip angle estimation, utilizing a data-driven approach that incorporates input from multi-modal sensors. This approach can be classified into two categories: pure AIA approach and hybrid AIA approach.

\subsubsection{Pure AIA approach} The pure AIA approach involves achieving vehicle sideslip angle estimation entirely through an end-to-end data-driven process, without relying on traditional mathematical models. Liu et al. develop a time-delay neural network to evaluate vehicle sideslip angle using input data such as steering angle, lateral acceleration, yaw rate, and wheel speed \cite{liu2020time}. Bonditto et al. construct three artificial neural networks to regress the sideslip angle in three road conditions - dry, wet, and icy. The estimated output is corrected by a road condition classifier network \cite{bonfitto2020combined}. Liu et al. propose a non-linear auto-regressive neural network to estimate vehicle sideslip angle. Through a comparative study, the proposed method shows strong performance across all driving conditions \cite{liu2020sideslip}.

\subsubsection{Hybrid AIA approach} The hybrid AIA approach, which combines kinematic or dynamical mathematical models with artificial intelligence techniques, is becoming increasingly popular for estimating vehicle sideslip angle. For instance, Kim et al. estimate vehicle sideslip angle and its uncertainty with long short-term memory (LSTM) aided by yaw rate, velocity, steering wheel angle, and lateral acceleration. The LSTM output is used as the input of the dynamic EKF/UKF model to enhance the performance of sideslip angle estimation \cite{kim2020vehicle}. Similarly, Novi et al. integrate an artificial neural network with UKF based on a kinematic model to estimate sideslip angle \cite{novi2019integrated}. Compared with the pure AIA approach, the hybrid approach can improve the interpretability of the estimation method to a certain extent.

The survey mentioned above highlights that estimating sideslip angle is not only critical for the trajectory tracking control of AVs but also essential for vehicle navigation, which is vital for communication among CAVs. Therefore, accurate sideslip angle estimation is indispensable for both AVs and CAVs.
\section{Trajectory Tracking control of AVs}

Vehicle trajectory tracking control is a crucial and fundamental component of AVs, enabling the computation of direct optimal commands like steering angles, throttle opening, and braking pedal to replace and assist the driver \cite{dominguez2016comparison, hu2019mme}. Within the control module, it can ensure that the vehicle accurately follows a predefined path while avoiding obstacles and maintaining safe driving conditions \cite{chen2022stochastic}. 
Accordingly, the feedback or self-learning tracking control algorithms at the microscopic level are designed to adjust the vehicle's speed, acceleration, and steering to follow the desired trajectory. Typically, the algorithms take into account the vehicle's dynamics and kinematics to determine the optimal control inputs. These inputs are adjusted in real-time to account for any uncertainties or disturbances in the vehicle's environment, such as changes in road conditions, traffic flow, or the presence of obstacles.

\begin{figure*}[ht]
\centering
\includegraphics[width=1\linewidth]{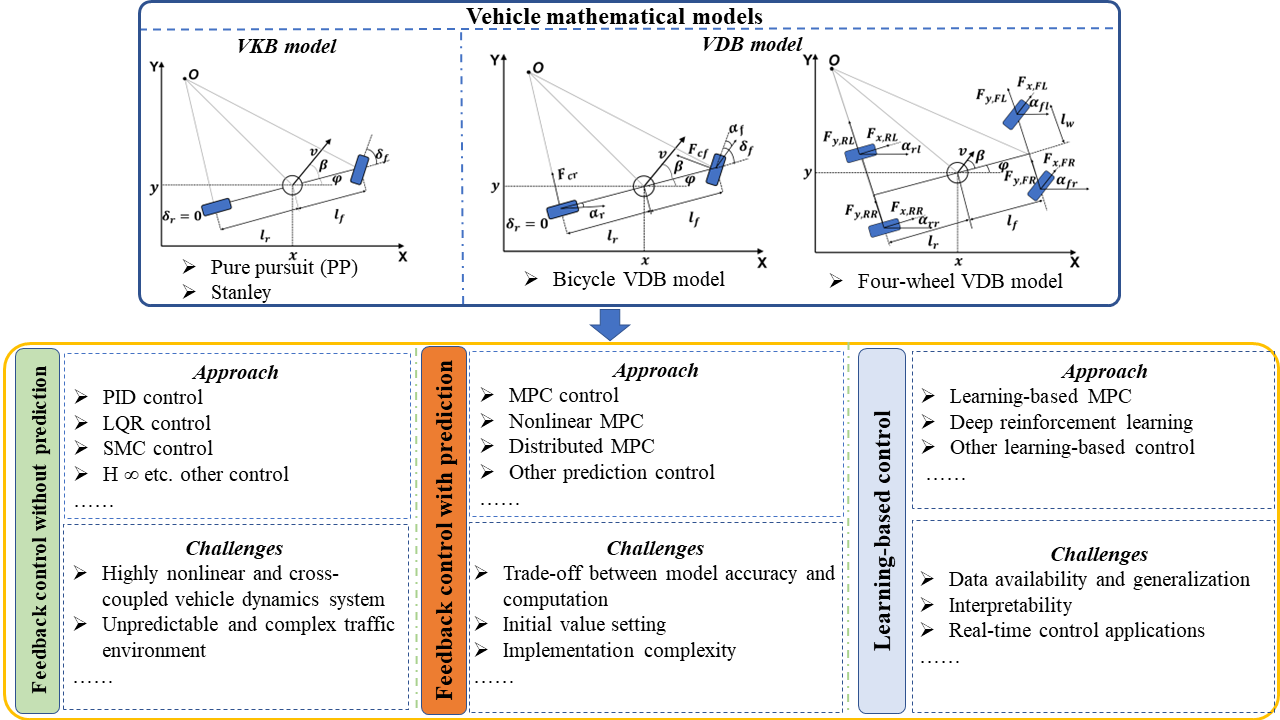}
\caption{{Summary of vehicle trajectory tracking control. VKB, vehicle-kinematics-based; VDB, vehicle-dynamics-based.}}
\label{fig_summary}
\end{figure*}

Nowadays, numerous institutions, automotive manufacturers, and component suppliers in the AV field have devoted significant attention to developing tracking control algorithms \cite{sun2022path, paden2016survey, kebbati2022lateral, zheng2019data, zhan2019trajectory}. These algorithms are primarily divided into the following three kinds: 1) feedback control without prediction (e.g., proportional-integral-derivative (PID), linear quadratic regulator (LQR), and sliding mode control (SMC) \cite{nagariya2020iterative, han2018energy, 7912351}); 2) feedback control with the prediction (e.g., model predictive control (MPC) \cite{8315037, hang2021active, wu2019fast, wang2020longitudinal}); and 3) learning-based control, such as the deep reinforcement learning. The overall framework is illustrated in Fig. \ref{fig_summary}.


\subsection{Vehicle mathematical model}
The trajectory tracking control of AVs involves the use of various vehicle models, which are typically classified into three types: VKB model \cite{kong2015kinematic, dominguez2016comparison}, bicycle VDB model \cite{chen2019dynamics, hua2019hierarchical, zhang2017nonlinear}, and four-wheel VDB model \cite{chen2019dynamics,xia2021advancing, hua2020research}. Different models have their own applicable scenarios \cite{hu2019rise}.

The VKB model is widely used as it could establish a relationship between the lateral/heading error and the steering angle directly. In general, the pure pursuit (PP) method is employed to calculate the lateral error from a fixed look-ahead distance on the desired path ahead of the vehicle. Typically, the reference point on the vehicle for this method is the center of the rear axle. The corresponding model is as follows:

\begin{equation}
    \begin{split}
        &\delta_f =\arctan(\frac{2Lsin\alpha }{L_d})\\
        &\kappa =\frac{2\sin \alpha }{L_d} \\
        &\sin \alpha =\frac{e}{L_d} \\
    \end{split}
\label{eq9}
\end{equation}
where $\delta_f$ is the steering angle; $L$ is the wheelbase; $\alpha$ is the angle between the vehicle’s body heading and the look-ahead direction; $\kappa$ is the curvature; $e$ is the lateral error; $ L_d $ is the look-ahead distance, which is a function of the vehicle velocity.

The equation above takes into account only the lateral error while disregarding the heading error, which is also crucial for ensuring optimal trajectory tracking performance. In order to address this issue, the Stanley method is developed, which considers both lateral and heading errors simultaneously. More specifically, the lateral error is defined as the distance between the closest point on the path and the front axle of the vehicle. The steering angle can be obtained using the following formula:

\begin{equation}
\delta_f =\arctan(\frac{ke}{1 + v(t)}) +\delta _{\psi} 
\label{eq10}
\end{equation}
where $e$ is the lateral error between the closest point on the path with the front axle of the vehicle; $ v(t)$ is the vehicle velocity; $k$ is the feedback gain; $\delta _{\psi}$ is the heading error. Compared with the PP method. more parameters need to be fine-tuned to ensure convinced performance.

Although the VKB model is widely used in vehicle trajectory tracking control, it ignores errors like tire slip which is pretty important for critical driving conditions. In order to reduce the above errors, many scholars also construct vehicle trajectory tracking control algorithms with the VDB model. Here is an example of a widely used bicycle VDB model. Its formula is as follows:

\begin{align}
    \begin{split}
    & \ddot{x} = \dot{\varphi }\dot{y} + a_x \\
    & \ddot{y} = -\dot{\varphi }\dot{x} + \frac{2}{m}({F_{cf}} cos \delta _f+F_{cr}) \\
    & \ddot{\varphi} = \frac{2}{I_z}(l_f{F_{cf}} - l_rF_{cr}) \\
    \end{split}
\label{eq11}
\end{align}


In order to further improve the accuracy of the vehicle model, some scholars also adopt the four-wheel VDB model \cite{guo2020computationally}. However, the improvement in accuracy will also lead to an increase in model complexity and computation. For this reason, the selection of the vehicle mathematical model ultimately depends on several factors, including driving scenarios, accuracy, and complexity.


\subsection{Feedback control algorithms without prediction}

Feedback control algorithms without prediction typically employ explicit control theories, such as PID, LQR, and SMC methods \cite{hu2019lane}. These techniques offer robust trajectory tracking performance under most driving conditions.

\subsubsection{PID control}

PID trajectory tracking control is a widely-used approach in industrial applications due to its simplicity and efficiency for AVs trajectory tracking control. Park et al. improve the trajectory tracking control accuracy of AVs by utilizing a proportional-integral (PI) control algorithm that minimizes the lateral distance error. The algorithm's proportional gain remains constant, while the integral gain varies based on the road curvature \cite{park2014development}. Similarly, Chen et al. employ a PI control algorithm along with a low-pass filter, based on the PP model, to enhance the trajectory tracking smoothness \cite{8569416}. In addition, Marino et al. utilize a nested PID steering control scheme and integrate the active steering function to perform trajectory tracking control for roads with uncertain curvature \cite{marino2011nested}. Mayyahi et al. employ a PID trajectory tracking control with parameters optimized by a PSO algorithm \cite{al2015path}. Han et al. propose an adaptive PID neural network trajectory tracking control strategy, where the model parameters are identified through the forgetting factor least squares algorithm, and the PID parameters are adjusted using backpropagation neural network \cite{han2017lateral}. To improve vehicle stability, vehicle yaw rate is also introduced into designing trajectory tracking control algorithms, which can improve the control performance \cite{bacha2018autonomous}. Besides VKB models, some scholars also adopt VDB models for AV trajectory tracking control.  Zhao et al. propose an adaptive-PID trajectory tracking control with a bicycle VDB model to ensure robust trajectory tracking control performance in the face of large-scale parameter variation and disturbances \cite{zhao2012design}.


In general, the PID controller is a relatively simple control algorithm to implement in AVs, and it can be easily programmed into microcontrollers or embedded systems to achieve industrial applications. However, PID trajectory tracking control tends to suffer from the difficulty of feedback gain tuning. When driving conditions change significantly in the real environment, the control parameters may no longer be optimal in some cases. Although adaptive PID control methods and self-learning PID control have been explored by many scholars, adaptive and automatic parameter tuning is still complicated and time-consuming. Furthermore, achieving satisfactory control performance using PID control under complex driving conditions remains a challenging task. \cite{kebbati2021optimized,8613388}.


\subsubsection{LQR control}
To alleviate the above limitations, some scholars also employ the LQR approach for vehicle tracking control. The objective of this approach is to minimize a quadratic cost function $J$ as in (\ref{lqr}). The feedback gain is solved by the corresponding Riccati equation.
\begin{equation}
argmin\kern0.1em J = \int_{0}^{\infty }{X}^T QX + {u}^T Rudt 
\label{lqr}
\end{equation}
where $Q$ is a positive semi-definite diagonal weight matrix; $R$ is a constant matrix penalizing the control effort; $X$ is the states matrix; $u$ is the control signal. As LQR is a state feedback control, $K$  is the feedback gain obtained by the variational method, as described by the following equation:
\begin{equation}
 K = {\left(R + {B}^T PB\right)}^{-1}{B}^T PA 
\label{eq13}
\end{equation}
where the matrix $P$ is the solution with the Riccati equation and given by:
\begin{equation}
{A}^EP + PA- PB^{-1}{B}^TP + Q = 0 
\label{eq14}
\end{equation}

Typically, the LQR approach is suitable for optimizing linear systems, where the parameters of Q and R need to be tuned to facilitate the system's convergence and stability. For vehicle trajectory tracking control, the LQR method is usually designed based on VDB models. Piao et al. present an LQR optimal trajectory tracking control with a bicycle VDB model, and they optimized the weight matrices of Q and R using a simultaneous perturbation random approximation algorithm \cite{piao2019lateral}. Choi et al. design the trajectory tracking control scheme by winding road disturbance compensator (WRDC) gain on a curved road from the dominant state for the lateral error in the lane-keeping control, and then the WRDC is applied to an LQR trajectory tracking control to compensate for the inaccuracy of linear system. Eventually, compared to the classical LQR methods, the proposed algorithm is demonstrated to be more effective \cite{choi2020vehicular}. Especially, vehicle trajectory tracking control under a critical driving situation is essential and should not be ignored in real applications \cite{xiong2021integrated}. Menhour et al. employ an LQR method with the LMI framework to solve the problem of switching steering control considering critical driving conditions \cite{menhour2014switched}. 
In some cases, the kinematic model is also utilized in the LQR method. Alcala et al. propose a non-linear kinematic Lyapunov-based control to address the problem of vehicle trajectory tracking control. The corresponding parameters are obtained through the LQR approach with a linear matrix inequality (LMI) expression form \cite{alcala2018autonomous}.


Based on the above survey, the LQR approach can provide an optimal control solution that minimizes the cost function, which can result in superior tracking performance through online or offline optimization. In particular, many researchers use the LQR method with LMI to achieve control convergence via typical Lyapunov function designs. However, LQR control is not suitable for systems with large control inputs because it assumes that the control input is small and does not cause significant deviations from the desired trajectory. Therefore, in some critical driving conditions, such as considerable road curvatures or low road adhesion, the trajectory tracking control performance may be compromised and result in poor robustness. Moreover, although LQR tracking control can handle nonlinear systems by linearizing the dynamics around the desired trajectory, linearizing the highly nonlinear and rapidly time-varying vehicle model is likely to result in model distortion. Therefore, more advanced techniques may need to be explored for the vehicle system.

\subsubsection{SMC control}
To address the issue of linearization assumptions in LQR, SMC is introduced in AVs trajectory tracking control. SMC is a robust control method for nonlinear systems with parametric uncertainties and external disturbances \cite{chen2019comprehensive}. The challenge in implementing SMC is to design a super-twisting plane to ensure robust stability and reduce the chattering phenomenon. The system function of SMC is as follows:
\begin{equation}
\dot{x} = f\left(t,x\right) + g\left(t,x\right)u(t)
\label{eq13}
\end{equation}
where $f$ and $g$ are continuous functions; $x$ and $u$ the state vector and the control input, respectively. Then a sliding variable $s$ with a derivative is expressed as follows:
\begin{equation}
\dot{s}\left(t,s\right) = \phi \left(t,s\right) + \Phi \left(t,s\right)u(t) 
\label{eq13}
\end{equation}
The sliding mode trajectory tracking control aims for the system to converge on the sliding surface, defined as $s = 0$. To achieve this goal, ${s}_0$, ${b}_{min}$, ${b}_{max}$, and ${C}_0$ should satisfy the following constraints:
\begin{equation}
\left\{\begin{array}{l}x\in {R^n}, \mid s\left(t,x\right)\mid < {s}_0\\ {}\mid u(t)\mid \leqslant {U}_{max}\\ {}0 < {b}_{min}\leqslant \mid \phi \left(t,s\right)\mid \leqslant {b}_{max}\\ {}\mid \phi \left(t,s\right)\mid < {C}_0\end{array}\right.
\label{eq13}
\end{equation}
Thus, the SMC can be given as follows:
\begin{equation}
\delta = u(t) = {u}_1 + {u}_2\left\{\begin{array}{l}{u}_1 = -\alpha \left |s \right | ^ {\tau}\mathit{\operatorname{sign}}(s),\tau \in \left[0,0.5\right]\\ {}{\dot{u}}_2 = -\beta sign(s)\end{array}\right. 
\label{eq13}
\end{equation}
Hence, the super-twisting algorithm based on a sliding surface yields the control signal of the steering angle $\delta $. Where $\alpha$ and $\beta$ are positive constants. And they should satisfy the following conditions:
\begin{equation}
\left\{\begin{array}{l}
\beta \geqslant \frac{C_{0}}{b_{\min }} \\
\alpha \geqslant \sqrt{\frac{4 C_{0}\left(b_{\max } \beta+C_{0}\right)}{b_{\min }^{2}\left(b_{\min } \beta-C_{0}\right)}}
\end{array}\right.
\end{equation}

For the trajectory tracking control algorithm of AVs, the lateral error is described by $e = y - {y_{ref}}$. And the desired lateral acceleration ${a}_{y_{ref}} = \frac{v^2}{R}$ can be expressed as a function of $R$, which is the radius of the road curvature. According to the VDB model, the error model is expressed as follows:
\begin{equation}
\ddot{e} = \ddot{y} + {v}\dot{\varphi}-\frac{v^2}{R}
\label{eq15}
\end{equation}

Due to its insensitivity to model uncertainties, SMC has become a popular nonlinear method for the trajectory tracking control of AVs, and the VDB model is widely exploited to describe vehicle motion. Akermi et al. employ a fuzzy system to automatically adjust the gain of SMC to compensate for variations in system parameters \cite{akermi2020novel}. Dai et al. propose to integrate SMC with particle swarm optimization (PSO) to improve the trajectory tracking control robustness of AVs. In detail, the authors design an SMC-based steering trajectory tracking control algorithm to eliminate lateral and heading errors \cite{dai2018force}. However, an inappropriate sliding surface may cause system oscillations. To eliminate this drawback, Li et al. present an adaptive SMC control scheme to improve handling and stability by reducing jittering. In this method, the weight of the sliding surface is adaptively scheduled according to the stability index \cite{li2021adaptive}.  Wang et al. design the high-order sliding mode to effectively reduce system chattering and improve control accuracy \cite{10025626}. As high-order sliding modes suffer from computation workload, the second-order sliding mode remains the mainstream to ensure the real-time performance of the algorithm. Alternatively, some scholars integrate SMC with other control methods to eliminate the jittering phenomenon. Zhang et al. present an optimal preview linear quadratic regulator (OPLQR) based on the SMC approach with a 2-DOF vehicle VDB model to enhance the smoothness of the vehicle motion after obtaining the desired steering angle \cite{zhang2019autonomous}. Nasr et al. achieve improved trajectory tracking performance by combining SMC and fuzzy logic control. From the experimental results, SMC has shown high robustness against external disturbances and sharp-edged paths, while fuzzy logic control can intelligently control maneuvers \cite{8519511}. Moreover, the four-wheel VDB model is also applied to achieve vehicle trajectory tracking control. Chen et al. propose the non-singular terminal SMC control algorithm to design a robust trajectory tracking control algorithm \cite{chen2019comprehensive}.


Overall, SMC is a practical control approach well-suited for nonlinear systems. This is because SMC is based on the concept of sliding mode, which refers to a condition where the process variable is forced to move along a predefined trajectory. Meanwhile, SMC is less dependent on the model parameters than other control techniques, making it more robust in the presence of model uncertainty and disturbances. Unfortunately, chattering near the sliding surface is still an issue, particularly in real-time applications. Additionally, SMC is more effective when the reference signal is smooth and continuous. If the reference signal is not smooth, such as the uneven road surface, it can result in high-frequency oscillations in the control input. Accordingly, a more complicated dynamics model should be designed by adding a degree of freedom. To address these limitations, researchers have also proposed hybrid control schemes that combine SMC with other control methods, such as PID or LQR.

\subsubsection{Other control methods}
Besides PID, LQR, and SMC, researchers have also applied some other control methods to optimize the trajectory tracking performance of AVs. These methods include H-infinity, adaptive, and fuzzy control \cite{9779983, xue2022nonlinear, an2021game, yu2019robust}. In H-infinity methods, the H-infinity norm of the system is minimized by solving an optimization problem involving the Riccati equation. Hu et al. present a robust H-infinity output-feedback control strategy using a hybrid genetic algorithm/linear matrix inequality approach that is robust to external disturbances and uncertainties in environmental parameters \cite{hu2016robust}. Zhou et al. propose an H-infinity lateral tracking control that explicitly accounts for the path-tracking kinematic nonlinearity and is demonstrated to be sector-bounded \cite{10.1115/1.4051466}. Zhou et al. also formulate a novel path-tracking control using the generalized H-infinity robust control method. Considering that tracking control problems typically involve noise, modeling uncertainties, and disturbances, adaptive control is also an effective solution for addressing nonlinear issues. For example, \cite{9693282} explicitly addresses the steering backlash issue with a novel adaptive backlash inverse compensator. Departing from the mainstream adaptive vehicle motion control design, \cite{9802699} tackles the trajectory tracking control problem by leveraging a novel non-quadratic Lyapunov design, which is demonstrated to have the potential to deliver superior transient performance. Wang et al. originate a driver-centered lane-keeping assist trajectory tracking control by unifying a non-certainty equivalent adaptive control scheme \cite{10037758}. Additionally, fuzzy control is also regarded as a feasible method for trajectory tracking control in AVs. Zhao et al. use fuzzy logic systems and neural networks to model human expertise through linguistic variables \cite{zhao2019identification}.


Although feedback control without prediction and other control algorithms have achieved remarkable progress in the past several years. The highly nonlinear and cross-coupled vehicle dynamics system, as well as the unpredictable and complex traffic environment, are still challenging for robust trajectory tracking control of AVs \cite{cheng2020robust}.

\subsection{Feedback control algorithms with prediction}

\begin{figure}[ht]
\centering
\includegraphics[width=1\linewidth]{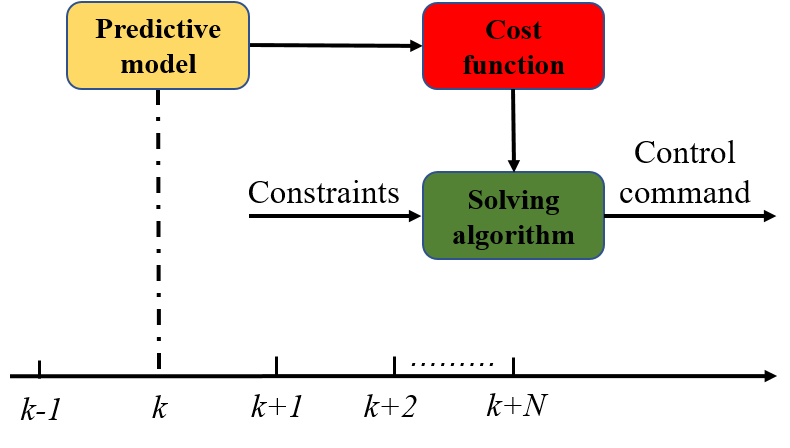}
\caption{{Block diagram of MPC.}}
\label{fig_mpc}
\end{figure}

MPC is a prominent approach to trajectory tracking control because it can solve optimization problems by introducing various constraints and prediction horizons \cite{guo2017nonlinear, xiang2021path, zhou2022interaction}. The general block diagram is shown in Figure \ref{fig_mpc}. Unlike LQR, MPC can predict future behavior using a prediction horizon and solve a finite-horizon open-loop optimal control problem. The optimal control sequence in a series of discrete time steps is computed by minimizing a cost function. The entire process is repeated iteratively, and only the first control input is applied to the vehicle system. Moreover, the MPC considers disturbances in the form of certain constraints to enhance the robustness and stability \cite{liu2021hierarchical, zhou2022interaction}. Thus, MPC always involves an intuitive parameterization by adjusting a process model at a higher computational effort cost than classical trajectory tracking control approaches.

The state space model of the MPC is presented as follows:

\begin{equation}
\begin{aligned}
& \boldsymbol{x}(\boldsymbol{k}+1) =\boldsymbol{f}(\boldsymbol{x}(\boldsymbol{k}) \boldsymbol{u}(\boldsymbol{k})) \\ 
& \boldsymbol{y}(\boldsymbol{k}) =\boldsymbol{h}(\boldsymbol{x}(\boldsymbol{k}))
\end{aligned}
\end{equation}

MPC minimizes a user-defined cost function $J$, e.g., the trajectory tracking error between the desired path  $\boldsymbol{r}$ and the model output $\boldsymbol{y}$.

\begin{equation}
    \begin{array}{cc}
    \min _{\boldsymbol{u}} & \boldsymbol{J}(\boldsymbol{x}(\boldsymbol{k}), \boldsymbol{u}(\cdot)) \\
    \min _{\boldsymbol{u}} & \sum_{i=N_1}^{N_2}\|\boldsymbol{r}(\boldsymbol{k}+i \mid \boldsymbol{k})-\boldsymbol{y}(\boldsymbol{k}+i   \mid \boldsymbol{k})\| \\
    \text { s.t. } & \boldsymbol{u}_{l b} \leq \boldsymbol{u}(\boldsymbol{k}+j \mid \boldsymbol{k}) \leq \boldsymbol{u}_{u b} \\
    & \boldsymbol{y}_{l b} \leq \boldsymbol{y}(\boldsymbol{k}+i \mid \boldsymbol{k}) \leq \boldsymbol{y}_{u b} \\
    & \forall i \in\left\{N_1, \cdots, N_2\right\} \text { and } j \in\left\{\left(0, \cdots, N_u\right\}\right.
    \end{array}
\end{equation}
where $\|\cdot\|$ is the arbitrary norm; ${N_{1}, \cdots, N_{u}}$ is the prediction horizon in a series of discrete time steps; $\boldsymbol{x}(\boldsymbol{k}+i \mid \boldsymbol{k})$ is the predicted state $\boldsymbol{k}+i$ at time point $\boldsymbol{k}$; $\boldsymbol{x}(\cdot)$ is a sequence of states. They can be described as follows:

\begin{equation}
    \begin{aligned}
    & \boldsymbol{x}(\boldsymbol{k}+i) \quad \forall i \in\left(0, \cdots, N_{2}\right) \Rightarrow \boldsymbol{x}(\cdot) \\
    & \boldsymbol{u}(\boldsymbol{k}+i) \forall i \in\left(0, \cdots, N_{u}\right) \Rightarrow \boldsymbol{u}(\cdot) \\
    & \boldsymbol{y}(\boldsymbol{k}+i) \forall i \in\left(N_{1}, \cdots, N_{2}\right) \Rightarrow \boldsymbol{y}(\cdot)
    \end{aligned}
\end{equation}

In this way, the constraint formulation will be abbreviated by $
\boldsymbol{x}_{l b} \leq \boldsymbol{x}(\cdot) \leq \boldsymbol{x}_{u b} \Rightarrow \boldsymbol{x} \in \mathbb{X}_{f}$, indicating that the sequence $\boldsymbol{x}(\cdot)$ is in the feasible set $\mathbb{X}_{f}$.

Several recent studies have demonstrated the exceptional performance of MPC for trajectory tracking control of AVs. They mainly focused on enhancing its performance in the following three key areas: improving the feasibility of open-loop optimization problems (e.g., by improving model accuracy), increasing the stability of closed-loop trajectory tracking control (e.g., through more accurate prediction and control processes), and enhancing robustness in the face of uncertainties (e.g., under extreme driving conditions). Regarding feasibility, Chu et al. propose a feedback MPC control that leverages dynamic and trajectory tracking characteristics to reduce the computing time and effort required for tuning parameters. By combining MPC with PID, they aim to minimize errors caused by the model's simplification. The experimental and simulation results demonstrate the effectiveness of the proposed method \cite{9715995}. Xu et al. develop a two-stage nonlinear model predictive control (NMPC) strategy for obstacle avoidance while following the center line under highway cruising conditions to improve model accuracy \cite{xu2019design}. Shen et al. also utilize NMPC to enhance optimization efficiency, primarily focusing on balancing computation cost and model accuracy \cite{shen2016modified}.
Regarding stability, Cheng et al. propose a lateral-stability-coordinated collision avoidance control system (LSCACS) based on MPC with different modes: a standard driving mode, a full auto brake mode, and a brake and stability mode \cite{8727723}. Zhang et al. develop a Gaussian process-based MPC with a classical two-layer structure to enhance stability control. More importantly, the corresponding stability proof is also conducted \cite{zhang2023vehicle}. Zhu et al. present a novel trajectory tracking control approach based on MPC to avoid high-frequency oscillations automatically by the switching algorithm. The uncertainty and external disturbances have been addressed and validated in the test vehicle \cite{zhu2017model}.
From the uncertainty perspective, Chen et al. design a hierarchical dynamic drifting trajectory tracking control under both drifting maneuvers and typical cornering maneuvers by blending MPC and LQR methods to achieve adaptably accurate trajectory tracking control 
 \cite{chen2023dynamic}. Moreover, the VKB model is also applied in MPC algorithms due to its simplicity. Zhang et al. design a trajectory tracking control based on the MPC using VKB model. The simulation results show that the test vehicle is capable of accurately following the desired path, even at sharp corners \cite{zhang2019trajectory}. Elbanhawi et al. propose an MPC based on a VKB model with PP to improve trajectory tracking performance at high speed \cite{elbanhawi2018receding}. Abdelmoniem et al. design the MPC trajectory tracking control method to address the restriction of sudden changes of heading angle and enhance the accuracy of heading error based on Stanley \cite{abdelmoniem2020path}.


MPC provides an optimal control solution that minimizes a performance criterion over a finite horizon, which can result in superior tracking performance compared to other control techniques. For another, MPC can handle nonlinear systems and constraints by using a nonlinear model of the system to predict future states and calculate the optimal control input  within safe limits. However, they still face several issues, such as deficiencies in the accurate model, prediction horizon limitations, and computational complexity, making it challenging to achieve real-time application \cite{xu2021custom, han2022coxhe, chen2023planning}. In addition, in real-world conditions, there exist signal delays from the sensors and actuators. However, MPC is sensitive to delays in the system, which can affect the prediction accuracy and compromise the control performance. Another challenge arises from setting the initial value in accelerating the optimization process. If the initial value is unsuitable, optimization may fail or require a prolonged period, with unpredictable calculation time for each step. 

\subsection{Learning-based control algorithms}
To address the above-mentioned challenges, there are several ways to optimize MPC control based on learning methods. One way is through sampling-based MPC algorithms that adopt simple policies to sample control sequences \cite{sacks2022learning, 9812369}. Another way is through self-learning model predictive control, which is closer to the optimal control rate based on known parameters \cite{chen2023lateral}. Additionally, there are learning-theoretic perspectives on MPC via competitive control that can help optimize the process \cite{shi2020online}. 

Furthermore, developing an accurate vehicle model is a challenging task due to the strong non-linearity and uncertainties present in AVs \cite{wang2018reinforcement, chen2020autonomous}. To overcome this challenge, learning-based control algorithms have been extensively explored as they are independent of a specific model and can address complicated nonlinear control systems \cite{xie2022modeling, hua2021surrogate, zhang2021online}. Deep reinforcement learning has emerged as a potential solution to the limitations of modern control algorithms for trajectory tracking control of AVs \cite{8986835, rizvi2018output, kuutti2020survey}.



\begin{figure}[ht]
\centering
\includegraphics[width=1\linewidth]{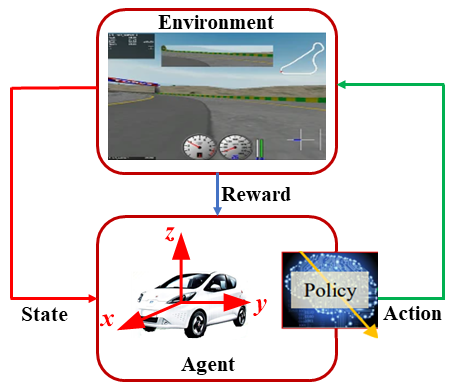}
\caption{{Block diagram of RL.}}
\label{fig4}
\end{figure}

When implementing a deep reinforcement learning algorithm, AVs learn an optimal control policy by interacting with the environment and utilizing gathered data, as shown in Figure \ref{fig4} \cite{yan2022multi, sutton2018reinforcement, hua2023energy, shuai2023optimal}. To overcome the challenges of existing trajectory tracking control algorithms, Zhao et al. propose a learning-based optimal control algorithm that approximates the critic and actor networks through two multi-layer neural networks \cite{zhao2017model}. Folkers et al. train a neural network agent using proximal policy optimization in a simulated environment to reach a specific target state while achieving trajectory tracking control \cite{8814124}. To reduce dependence on specific approximation structures, which is a limitation of deep reinforcement learning algorithms, Zhu et al. propose a probably approximately correct (PAC) algorithm that can efficiently utilize online data \cite{zhu2015data}. Wulfmeier et al. present maximum entropy deep inverse reinforcement learning (MEDIRL), a framework that applies high-capacity neural network architectures and extends scalability concerning the complexity of the environment, behavior, and size of training datasets \cite{wulfmeier2017large}. Zhang et al. propose a novel method for AVs trajectory tracking control, optimizing the residual policy with reinforcement learning algorithms on the basis of the guiding policy from a designed modified artificial potential field controller. Extensive experiments illustrate the method outperforms the leading algorithms \cite{zhang2022residual}. 
Moreover, to achieve optimality and efficiency in various driving circumstances, Chen et al. combine the proximal policy optimization (PPO) algorithm with a PP mathematical model. The PP model generates a baseline steering control command, while the PPO derives a correction command to improve trajectory tracking performance. The combination of the two controllers results in a more robust and adaptive operation \cite{chen2021deep}.

Admittedly, deep reinforcement learning algorithms also have limitations, including the need for extensive training data, the potential for over-fitting, and the challenge of interpreting the model's inner workings. Despite the existing challenges, the field of deep reinforcement learning is rapidly expanding with new techniques and algorithms that aim to enhance its capabilities.

In addition to the learning methods mentioned above, Liu et al. introduce a novel learning-based adaptive control (MFAC) algorithm. This algorithm is based on the dual successive projection (DuSP)-MFAC method, which utilizes the newly introduced DuSP approach and the symmetrically similar structures of the trajectory tracking control and parameter estimator of MFAC. By utilizing the preview-deviation-yaw angle, the trajectory tracking problem is transformed into a stabilization problem \cite{8641438}. Wang et al. develop a novel adaptive data-driven vehicle trajectory tracking control system by innovatively combining extremum seeking and learning-based control \cite{9714714}. Sharma et al. train two distinct neural networks to predict vehicle speed and steering based on the road trajectory using end-to-end learning in the open racing car simulator (TORCS). The results show that the trajectory tracking control performed effectively on two tracks despite the limited training data available \cite{sharma2019lateral}.


Overall, vehicle trajectory tracking control algorithms in AVs have been reviewed in depth at the microscopic level. The selection of the appropriate algorithm depends on the specific application and performance requirements of the vehicle trajectory tracking control \cite{chen2020neuroiv, lu2021monet}.



\section{Collaborative Control of CAVs}
AVs equipped with sensors such as cameras, radar, and LiDAR are a critical component of ITS \cite{lu2021hregnet, zhan2019computational, li2022egocentric, liu2020globally, hu2022robustness, hu2022investigating, li2023voxformer, hu2021domain, hu2019retrieval}. However, despite their advanced sensor technology, it is not possible for AVs to fully and reliably perceive the dynamic and variable environment all the time with only their onboard sensors for example due to adverse weather conditions, sensor or vehicle model uncertainties, and illumination variation \cite{liu2020globally}.

Fortunately, the combination of the IoT and AI has resulted in significant advancements in ITS \cite{wu2022continual, xu2022opv2v, li2022v2x, chen2020event}. These technologies enable interconnectivity among various traffic participants, including vehicles, traffic signals, humans, and infrastructure, to enhance transportation efficiency, reduce emissions, and prevent accidents. The collaboration among these agents, in particular, CAVs, through V2X communication can enhance their decision-making capabilities and make the transportation system more efficient and sustainable \cite{dong2021space, dong2020spatio}. To this end, to address the limitations of AVs, collaborative control for CAVs has emerged as a promising technique in ITS. Collaborative control techniques involve sharing information beyond the line-of-sight and field-of-view to achieve coordinated decision-making among CAVs, thereby reducing the likelihood of accidents and improving overall traffic efficiency, comfort, and economy in typical traffic scenarios \cite{guan2022discrete, liu2020synchronization, li2021learning}. This technique has gained considerable interest due to its ability to optimize vehicle control ability on a relatively large scale with other CAVs. 

\begin{figure*}[t]
\centerline{\includegraphics[width=1.0\linewidth]{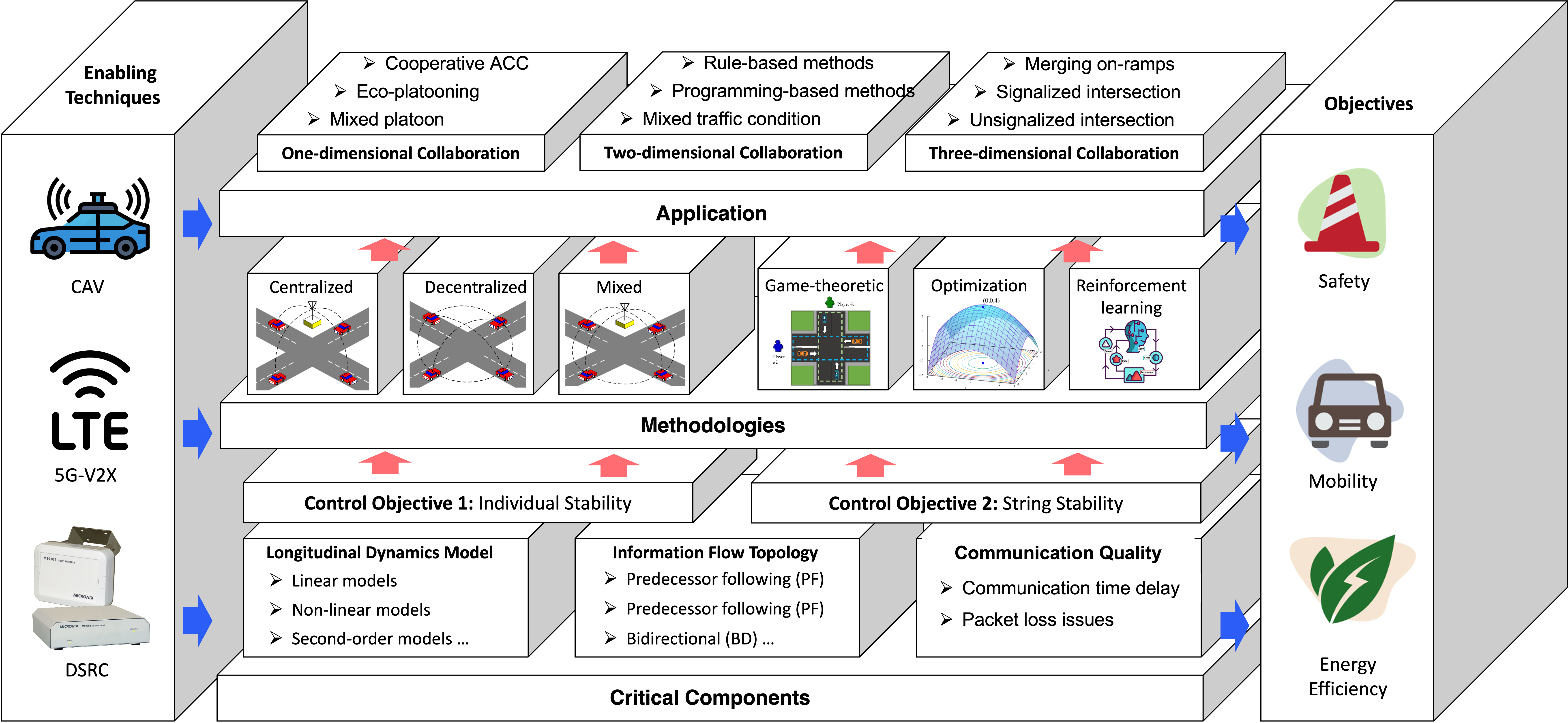}}
\caption{Summary of collaborative control techniques for CAVs and their applications. CAV, connected automated vehicle; DSRC, dedicated short-range communications.}
\label{zhiyun-sum}
\end{figure*}

In this section, we explored the enabling techniques, critical components of collaborative control techniques, collaborative control methodologies, and potential applications of the collaborative control technique for CAVs. The framework of this section is illustrated in Fig. \ref{zhiyun-sum}.

\subsection{Enabling techniques}

Collaborative control of CAVs relies on V2V and V2I to enable multiple CAVs to coordinate their movements and share information about their speed, position, and destination. Some advanced systems also share visual information, including detected bounding boxes~\cite{chen2022model} and intermediate neural features~\cite{xu2022bridging}. This technology enables smoother traffic flow, reduced congestion, and improved road safety, by facilitating the synchronization of acceleration, braking, and turning, among other factors. 
One of the key benefits of collaborative control is the potential to optimize traffic flow by reducing the need for braking and accelerating, resulting in fuel savings, reduced emissions, and improved energy efficiency. 
Another benefit is the reduction of traffic jams and collisions, which can lead to improved travel times, reduced vehicle wear and tear, and lower operating costs. 
For instance, by leveraging the shared visual information from multiple vehicles and infrastructure sensors, object detection accuracy can increase by more than 20\% \cite{xu2022v2x}. Cai et al. have demonstrated that optimizing the infrastructure's positioning can further increase this benefit through more realistic simulated LiDAR sensors \cite{cai2022analyzing}. Additionally, Xu et al. found that cooperation can still be highly beneficial even when the deployed vision models between vehicles and smart infrastructure differ, with the application of domain adaptation techniques \cite{xu2022bridging}. Furthermore, CAVs could also provide enhanced mobility services to users, such as adaptive routing and ride-sharing.  

\subsection{Critical components of collaborative control techniques}

\subsubsection{Description of collaborative control systems}
This section presents brief descriptions of the critical components of collaborative control systems.

\paragraph{Longitudinal dynamics model}
The longitudinal dynamics model is crucial for the collaborative control of CAVs. It predicts how CAV's speed and acceleration change based on various factors, such as dynamics, driving behavior, road conditions, and traffic flow. This model improves driving behavior, leading to more efficient driving, and is used for platooning, merging, and lane-changing.

There are various types of longitudinal dynamics models that differ in complexity and accuracy. The choice of dynamics model depends on the required accuracy, computational resources, and data availability. 
For example, linear models are the simplest, assuming a linear relationship between acceleration and inputs such as throttle position, braking force, and external disturbances. 
Non-linear models \cite{5876300} offer greater accuracy and the ability to handle more complex situations. They are mathematically derived from the dynamics of the vehicle and driving behavior. Second-order models \cite{5571043} consider acceleration and velocity, while third-order models add position to these variables. Higher-order models provide a more accurate representation of the vehicle's behavior but require more computational resources and data.
In addition, general linear models \cite{845117} are versatile, adaptable, and can represent a range of driving scenarios. They use linear equations to describe the relationship between inputs and outputs such as speed and acceleration. Statistical methods and data-driven approaches determine the coefficients, enabling the model to learn from real-world driving data.

\paragraph{Information flow topology}
The concept of information flow topology in the context of collaborative control of CAVs refers to the way information is exchanged between multiple vehicles to achieve coordinated and efficient movement. This is essential to ensure that the vehicles move together in a safe and efficient manner, reducing the likelihood of accidents and improving the overall flow of traffic.

There are several classification methods for information flow topology, each with its unique approach to exchanging information and achieving coordinated movement. 
These methods include predecessor following (PF) \cite{5571043}, predecessor leader following (PLF) \cite{swaroop1999constant}, bidirectional (BD) \cite{knorn2014passivity}, bidirectional leader (BDL) \cite{zheng2015stability}, two-predecessor following (TPF) \cite{swaroop1999constant}, and two-predecessor-leader following (TPLF) \cite{li2015overview}, among others. 
Each method varies in its level of coordination and flexibility, with some allowing for more advanced coordination and others allowing for more flexibility in movement.
The choice of method depends on the specific context and goals of the group movement, such as minimizing delays, reducing fuel consumption, or preventing accidents. Overall, the concept and classification methods of information flow topology are critical to the successful implementation of collaborative control of CAVs, as they enable safe and efficient movement of multiple vehicles on the road.


\paragraph{Communication quality}

In collaborative control of CAVs, effective communication is essential for safe and efficient operation. However, the communication quality issues can arise when multiple vehicles communicate with each other, leading to delays or loss of data. These can be caused by factors such as network congestion, signal interference, or hardware failures.

For instance, one significant issue that can arise is communication time delay \cite{di2014distributed,qin2016stability}. This refers to the time taken for data to be transmitted from one vehicle to another and can be caused by various factors such as limited bandwidth, transmission errors, or network congestion. Delays in communication can result in a loss of coordination between vehicles, leading to accidents or collisions. To mitigate this issue, techniques such as message prioritization \cite{selvi2018efficient}, efficient routing algorithms \cite{magaia2022group}, and congestion control methods \cite{gomez2016dependability} can be used.

Another critical issue is packet loss \cite{ploeg2014graceful}. This occurs when data packets are lost during transmission and can be caused by factors such as network congestion, signal interference, or hardware failures. Packet loss can lead to incorrect or incomplete information being received by other vehicles, resulting in unsafe or inefficient operation of the system. Therefore, techniques such as packet retransmission \cite{asefi2012mobility}, error correction codes \cite{studer2009flexible}, and network redundancy \cite{dietzel2016resilient} can be used to mitigate it.

\subsubsection{Objectives of collaborative control systems}
The collaborative control of an autonomous platoon aims to ensure all the vehicles in the same group move at a consensual speed while maintaining the desired spaces between adjacent vehicles, and thus increase traffic capacity, improve traffic safety, and reduce fuel consumption. The stability properties of the platoon system are the foundation of all the above-mentioned control objectives. 
The control objectives for collaborative control techniques can be broadly categorized into individual stability and string stability.

\paragraph{Individual stability}

Individual stability refers to the stability of each CAV in the system \cite{zheng2015stability}. The individual stability objective is achieved by ensuring that each CAV maintains its desired speed and distance from neighboring vehicles \cite{dolk2017event}. The control algorithm should be designed to prevent overshooting or undershooting the desired speed and distance while accounting for noise and disturbances. Achieving individual stability is essential because it provides a basis for ensuring the overall stability of the system.

To maintain individual stability, each CAV must be equipped with control systems that can accurately measure its speed and position relative to the other vehicles in the platoon. The control system must also be able to adjust the CAV's speed and acceleration to maintain the desired position and speed within the platoon. This can be achieved through the use of advanced sensing and control algorithms, such as adaptive cruise control (ACC) and cooperative adaptive cruise control (CACC).

\paragraph{String stability}

String stability, on the other hand, refers to the stability of the entire platoon or string of CAVs in the system \cite{jin2014dynamics}. String stability is achieved by ensuring that the CAVs in the platoon maintain a constant and safe inter-vehicle distance while traveling at a constant speed \cite{ploeg2013lp}. The control algorithm should ensure that any disturbances or fluctuations in speed or distance are quickly compensated for, and the platoon returns to its desired state. 

Maintaining string stability is more challenging than individual stability, as it requires the coordination of multiple vehicles with different dynamics. To maintain string stability, the control system must ensure that each CAV in the platoon follows a predefined spacing policy that dictates the distance between each vehicle in the platoon and how that distance is maintained during operation. V2V communication can be used to exchange information on the position and speed of each CAV and adjust the spacing policy accordingly \cite{guo2022distributed}. The control system must also be able to detect any disturbances or disruptions in the platoon and respond appropriately to maintain the desired spacing policy.

Several approaches can be used to maintain string stability, including consensus-based control \cite{li2018consensus}, MPC \cite{dunbar2011distributed}, feedback control \cite{orosz2016connected}, and feedforward control \cite{li2019parsimonious}. 
Consensus-based control involves designing a distributed control algorithm that enables each CAV to adjust its speed and position based on information from neighboring vehicles. 
MPC utilizes a dynamic model of the system to predict its future behavior and optimize its control inputs accordingly. 
Feedback control adjusts the speed and position of each vehicle based on sensor feedback and communication with other vehicles in the platoon. 
Feedforward control predicts the control inputs required to maintain string stability based on the current state of each vehicle and the desired trajectory of the platoon.

\subsection{Collaborative control methodologies}

\subsubsection{Coordination scheme}
The coordination scheme for multiple vehicles is a critical component of collaborative control for CAVs, which enables them to work together efficiently and safely. This scheme refers to the methods and techniques used to coordinate the behavior of multiple CAVs to achieve a common goal.
There are different coordination schemes for multiple vehicles, and they can be classified based on the control mode they use. The three main control modes are centralized, decentralized, and mixed control modes.

\paragraph{Centralized control}
Centralized control mode involves a central authority, such as a control tower or a traffic management system, controlling all vehicles. The central authority collects traffic information and uses it to optimize vehicle behavior. Although this mode provides excellent coordination and optimization, it requires a high level of communication and computational power.
\paragraph{Decentralized control}
In contrast, in the decentralized control mode, each CAV is autonomous and makes its own decisions based on local information. Each vehicle communicates with its neighbors and adjusts its behavior to avoid collisions and maintain the desired formation. This mode is more scalable and resilient than centralized control, but it may not provide optimal coordination. 
\paragraph{Mixed control}
In addition, the mixed control mode is a combination of centralized and decentralized control modes. In this mode, some aspects of the coordination are handled centrally, while others are handled locally. For example, a central authority can provide high-level commands, such as the desired route and speed, while each vehicle adapts its behavior to local conditions. This mode allows for a flexible and adaptive coordination approach that can achieve both global optimization and local responsiveness.

Each coordination scheme has its advantages and disadvantages, and the selection of the appropriate scheme depends on the specific application requirements and the available resources. Centralized control is more suitable for highly autonomous vehicles operating in a well-defined environment, such as a closed campus or a dedicated lane on a highway. 
The decentralized control is more suitable for low-level autonomy, such as platooning, where a group of CAVs follows a lead vehicle. 
Mixed control is suitable for applications that require both global coordination and local adaptation, such as urban traffic management.  
By classifying the coordination scheme based on the control mode, the optimal approach can be chosen depending on the specific application and the level of autonomy of the CAVs.

Collaborative control techniques optimize the behavior of multiple vehicles by coordinating their actions. Unlike traditional control methods for individual vehicles, collaborative techniques rely on inter-vehicle communication to exchange information about their states and objectives. They must consider the complex interactions and dependencies between multiple vehicles, and be able to adapt to changing conditions in real-time. However, designing these methods is challenging due to the significant number of variables to consider and the need for robust algorithms that can handle unexpected situations.

\subsubsection{Methodologies}

Collaborative control techniques leverage advanced algorithms and computational methods to enable CAVs to communicate and cooperate with each other in real-time. Collaborative control methodologies can be categorized based on the mathematical model used to design the control algorithms, including game-theoretic, optimization-based, and reinforcement learning-based methods. In the following sections, we will provide a detailed introduction to these three methods and their applications in collaborative control for CAVs.

\paragraph{Game-theoretic methods}

Game-theoretic methods model the interactions between multiple CAVs as a game, where each vehicle is a player aiming to optimize its own objective while considering the actions of other players. The control algorithms are then formulated to identify Nash equilibrium, where no player can increase their outcome by unilaterally changing their actions. Game-theoretic methods are beneficial in managing complex and dynamic interactions between multiple vehicles. 

For example, Liao et al. propose a cooperative driving game for CAVs in which each vehicle decides its optimal speed to minimize energy consumption while avoiding collisions with other vehicles \cite{liao2021game}. The game-theoretic approach ensures that each vehicle converges to a Nash equilibrium where all players optimize their objective functions simultaneously. 
Additionally, Rahmati et al. propose a game-theoretic framework for CAVs, in which vehicles cooperated to optimize their travel time by adjusting their speeds and routes \cite{rahmati2021helping}. The authors use a Stackelberg game model to capture the interactions between vehicles, where a leader vehicle set the rules for other followers to follow.
To further enhance the effectiveness of game-theoretic methods, Gong et al. propose a game-theoretic reinforcement learning approach for CAVs, where vehicles learn to cooperate to maximize the overall system performance \cite{gong2022modeling}.
However, it is still computationally expensive and challenging to implement in practice.

\paragraph{Optimization-based methods}

Optimization-based methods formulate the collaborative control problem as an optimization problem, where the objective is to find the optimal actions for each vehicle that maximize the overall performance of the group \cite{yang2021cooperative}. The algorithms can be designed using linear \cite{wu2021cooperative,liu2022single} or non-linear \cite{tajalli2021traffic} optimization techniques and are adaptable to handle a broad range of scenarios. 

Among this scheme, various programming models are widely used by researchers to pursue the global-optimal traffic operation scheme. For example, 
You et al. propose a mixed-integer linear programming (MILP) model to collaboratively optimize the trajectories of CAVs in terms of total vehicle delay, considering car-following and lane-changing behaviors \cite{yu2019corridor}.
In addition, Tajalli et al. propose a methodology that uses a mixed-integer non-linear program to optimize signal timing and trajectory control at intersections with a mix of CAVs and human-driven vehicles \cite{tajalli2021traffic}. 

In general, optimization-based methods may not be equipped to handle the uncertainties and unpredictability of real-world environments \cite{guan2020centralized,chen2020hierarchical}.


\paragraph{Reinforcement learning-based methods}

Reinforcement learning-based methods entail training the control algorithms using a trial-and-error approach, where the algorithms learn from their own experiences in the environment \cite{harada2022behavior}. The algorithms can be designed using deep learning techniques and can adapt to changing environments.

In order to utilize the advantages of reinforcement learning, various researchers have trained their driving agents which have the ability to makes decisions based on their own observation. For example, Chen et al. propose an intelligent speed control approach using deep reinforcement learning for CAVs with the purpose of improving safety, efficiency, and ride comfort \cite{chen2023safe}.
Additionally, Valiente et al. propose a decentralized framework for training CAVs to operate with human-driven vehicles by formulating the mixed-autonomy problem as a multi-agent reinforcement learning (MARL) problem. They optimize for social utility while prioritizing safety and adaptability \cite{valiente2022robustness}.
Furthermore, Raja et al. propose a block-chain integrated multi-agent reinforcement Learning (BlockMARL) architecture to enhance the efficiency of CACC while collaboratively detecting attacks and securely notifying the overall network \cite{raja2022blockchain}.

 However, reinforcement learning-based methods may require large amounts of training data and can be difficult to interpret and validate \cite{chen2021graph}.

\subsection{Applications}
In order to develop algorithms capable of real-life applicability, most of the existing research about collaborative control techniques focuses on specific driving scenarios or traffic facilities, such as platooning, lane-change, merging, and intersection.
This section will provide an overview of some of the most promising applications of collaborative control techniques for CAVs and their potential benefits and challenges.
This paper categorizes these applications into three groups according to the level of collaboration, namely one-, two-, and three-dimensional collaboration. Each level of collaboration presents unique technical and operational challenges, but also offers opportunities for improving safety, efficiency, and comfort in transportation \cite{xu2021opencda,xu2023opencda}.


\begin{figure}[b]
\centerline{\includegraphics[width=0.9\columnwidth]{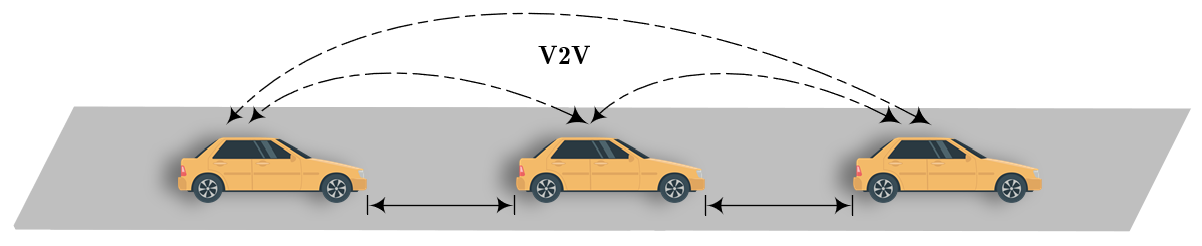}}
\caption{Illustration of the collaborative platooning at highways. 
}
\label{zhiyun-platoon}
\end{figure}

\subsubsection{One-dimensional collaboration}

One-dimensional collaboration involves coordinating the longitudinal (i.e., forward/backward) motion of vehicles, such as in platooning. 

\paragraph{Platooning}
In a platoon, collaborative control technology enables CAVs to travel in a short inter-vehicle distance to each other, with the lead vehicle dictating the speed and route for the other vehicles in the platoon. This approach can significantly improve fuel efficiency and reduce emissions, by minimizing the drag and air resistance between vehicles \cite{wang2017developing, wang2022gaussian}. Additionally, platooning can help reduce congestion and improve traffic flow, especially in urban areas with high levels of traffic density. 


This technique has been studied extensively for both trucks \cite{tsugawa2016review} and personal vehicles \cite{wang2019survey,axelsson2016safety}, with promising results. 
For example, Sun et al. demonstrate that truck platooning can reduce fuel consumption by about 10\% on highways \cite{sun2021investigating}, depending on the distance between the vehicles and the speed. In addition to energy savings, platooning can also enhance safety on the road. The use of advanced sensors and communication technologies enables the platoon to brake and accelerate in unison \cite{van2017evaluation}, reducing the risk of accidents caused by sudden changes in traffic flow.

In order to deal with the control problem in the platoon formation phase, Saeednia et al. \cite{saeednia2016consensus} design consensus protocols for vehicles such that their motion states converge to a common desired value. However, such heuristics cannot yield optimal solutions since their optimization processes are based on pre-defined motion patterns. Therefore, Deng et al. present an evolutionary algorithm to optimize the longitudinal trajectories of multiple vehicles with an energy-aware objective function \cite{deng2022coevolutionary,deng2022longitudinal}, which does not assume the underlying solution landscape compared with heuristics and require relatively shorter runtime than programming models.

\subsubsection{Two-dimensional collaboration}
Two-dimensional collaboration involves coordinating the lateral (i.e., left/right) motion of vehicles, such as in lane changing. 
\paragraph{Lane-changing}
With the support of collaborative control techniques, the safety and efficiency of lane changes can be improved even in situations with limited visibility or at high speeds, by enabling vehicles to communicate with each other and coordinate their movements. In addition, this technology can anticipate and prevent collisions by detecting and avoiding potential hazards automatically.



To optimize lane changing decisions and improve traffic flow, researchers have proposed various collaborative driving strategies. For instance, Lin et al. present a collaborative lane changing strategy using transferable utility games framework that improves individual and social benefits without adverse effects on traffic conditions \cite{lin2019pay}. Similarly, Ali et al. develop a game theory-based mandatory lane-changing model for traditional and connected environments. \cite{ali2019game}.

Other researchers propose optimal control policies for the ego CAV to implement a lane change maneuver in cooperation with neighboring CAVs. 
Chen et al. present such policies, optimizing both maneuver time and energy consumption while ensuring safety constraints \cite{chen2020cooperative}.
Tajalli et al. propose a mixed-integer program methodology for optimal control of CAVs in freeway segments with a lane drop, using a collaborative distributed algorithm to coordinate lane-changing decisions and improve traffic flow and capacity \cite{tajalli2022distributed}.
Furthermore, Xu et al. present a bi-level framework for collaborative driving of CAVs in conflict areas. They combine Monte Carlo Tree Search with heuristic rules to provide near-optimal solutions, resulting in improved traffic performance with consideration of both critical conflict zones and lane change right-of-way \cite{xu2020bi}. 

\begin{figure}[t]
\centerline{\includegraphics[width=0.9\columnwidth]{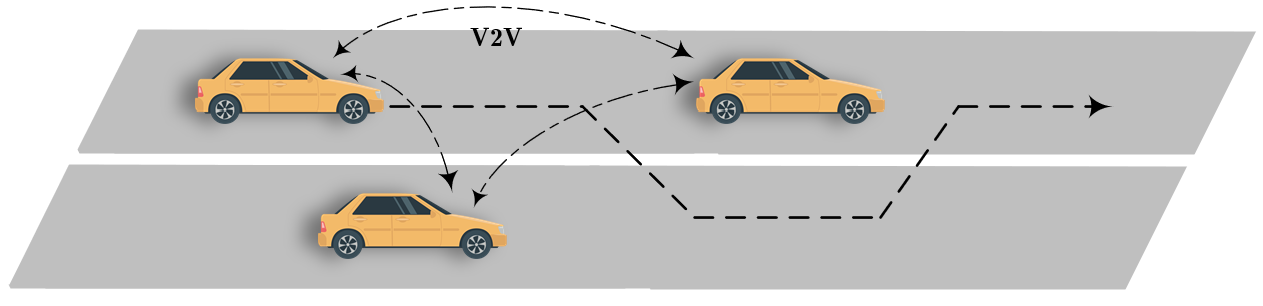}}
\caption{Illustration of the collaborative lane-change scenario.
}
\label{zhiyun-change}
\end{figure}

\begin{figure}[t]
\centerline{\includegraphics[width=0.9\columnwidth]{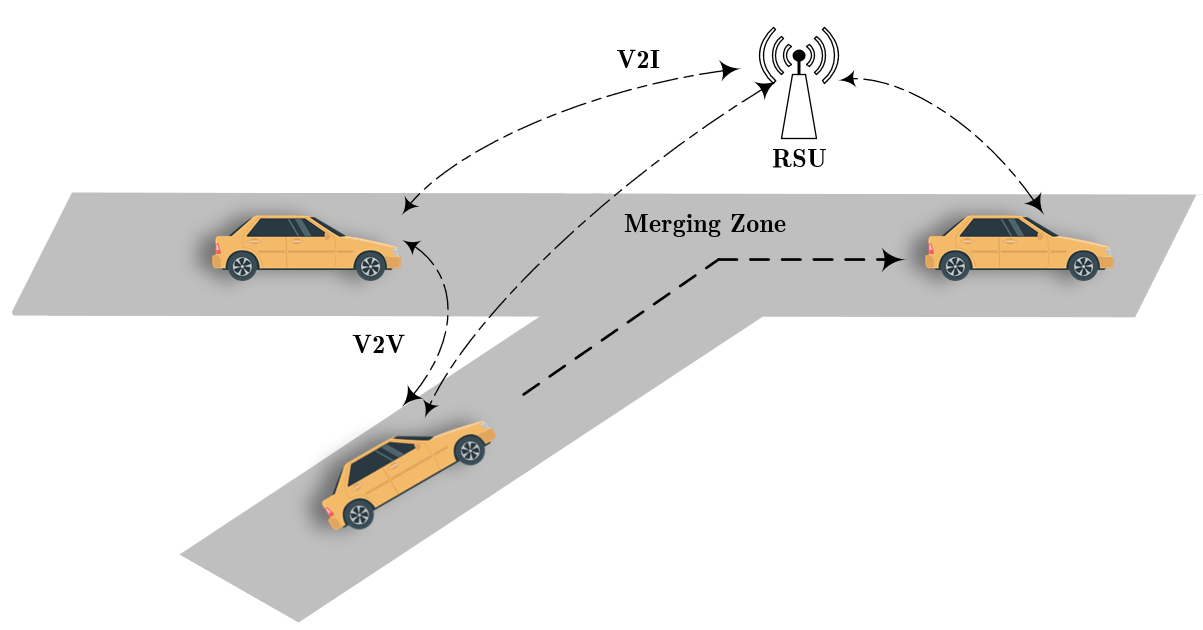}}
\caption{Illustration of the collaborative driving scenario at merging points.
}
\label{zhiyun-merge}
\end{figure}





\subsubsection{Three-dimensional collaboration}

Three-dimensional collaboration involves coordinating both longitudinal and lateral motion, as well as the time dimension, particularly in scenarios where multiple vehicles are approaching an intersection or merging point from different directions. To avoid collisions, vehicles must coordinate not only their longitudinal and lateral motion but also their timing and entry sequence, which requires advanced communication, sensing, and decision-making capabilities.

\paragraph{Merging on-ramps}
When it comes to merging at highway on-ramps, collaborative control techniques can help reduce congestion and improve traffic flow by enabling vehicles to safely and efficiently merge into traffic. By sharing information about their position and speed, vehicles can coordinate their movements to avoid collisions and smoothly merge with other traffic on the highway \cite{williams2021position}. This approach can also reduce the need for sudden braking and acceleration, leading to improved fuel efficiency and reduced emissions.

To achieve collaborative merging, Ding et al. present a rule-based adjusting algorithm for coordinating the merging of two strings of vehicles at highway on-ramps efficiently and safely in the longitudinal direction and evaluates its performance through simulation-based case studies under both balanced and unbalanced scenarios \cite{ding2019rule}.
Moreover, Liao et al.  propose a collaborative ramp merging system for CAVs using a digital twin approach based on vehicle-to-cloud communication that provides advisory information to improve safety and environmental sustainability during merging with an acceptable communication delay, as demonstrated in a real-world field implementation in Riverside, California \cite{liao2021cooperative}.

In order to reduce the complexity of problem-solving, 
Jing et al. present a hierarchical and decentralized collaborative coordination framework for the integrated longitudinal and lateral control of CAVs approaching on-ramps, which optimizes fuel consumption and passenger comfort, reduces fuel consumption, and improves traffic efficiency compared to baseline \cite{jing2022integrated}.
In addition, game theory has been widely used by researchers to model the collaborative control process. For example, Chen et al. propose a collaborative merging strategy for CAVs based on collaborative game theory and optimal control to improve traffic efficiency, reduce fuel consumption and enhance driving comfort through the construction of an economic payoff function that determines the merging sequence and coordinated merging trajectory \cite{chen2022cooperative}.


For the scenarios of mixed traffic, Sun et al. study collaborative control method for CAVs and conventional human-driven vehicles by developing a bi-level optimization program that guarantees system-efficient solutions and results in smoother ramp merging with an increase in traffic throughput of 10-15\% \cite{sun2020cooperative}.
Karimi et al.  outlines a hierarchical control framework for merging areas in mixed traffic composed of CAVs and human-driven vehicles, with a focus on the lower level control algorithm that establishes a set of collaborative CAV trajectory optimization algorithms for different merging scenarios through the use of model predictive control \cite{karimi2020cooperative}.

\begin{figure}[t]
\centerline{\includegraphics[width=0.9\columnwidth]{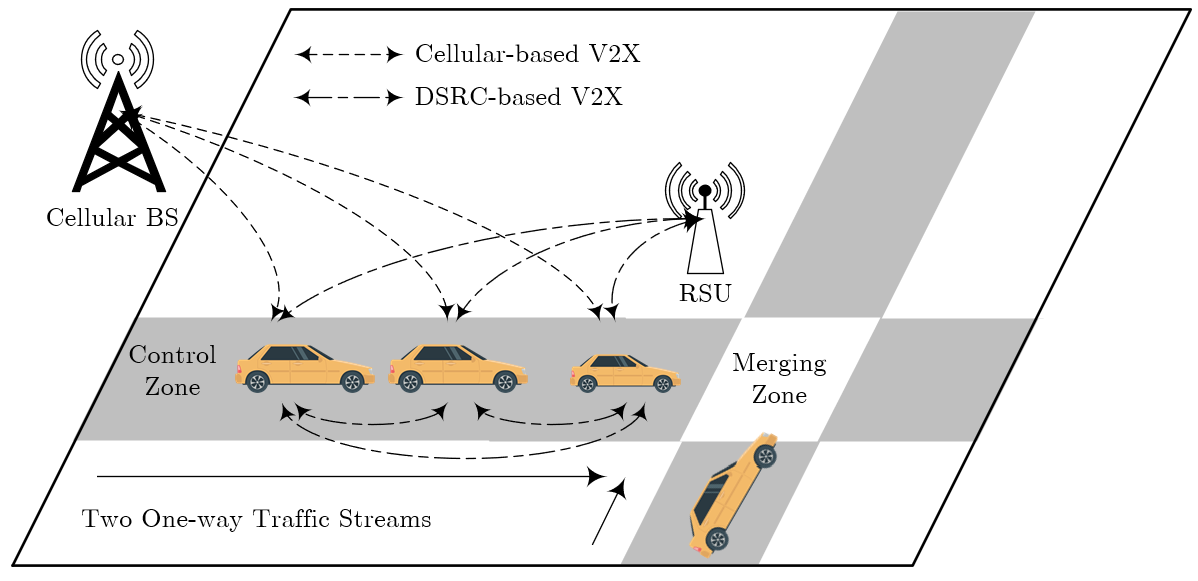}}
\caption{Illustration of the collaborative driving scenario at intersections \cite{deng2023cooperative}.
}
\label{zhiyun-intersection}
\end{figure}
\paragraph{Signalized intersections}


In signalized intersections, collaborative control technology can be used to enable vehicles to communicate with each other and the infrastructure to optimize traffic flow and reduce the likelihood of accidents \cite{zhou2022cooperative}. By enabling vehicles to anticipate and respond to traffic signals, collaborative control technology can help to reduce wait times and congestion, improve the safety of turning movements, and enable more efficient use of intersection capacity \cite{zhao2018platoon,feng2019composite,wang2019cooperative,du2021coupled}.
For example, Wang et al. propose a collaborative eco-driving system for signalized corridors that uses a role transition protocol for CAVs, leading to a reduction in energy consumption and pollutant emissions as CAV penetration rate increases \cite{wang2019cooperative}.
Liu et al. present an approach that optimizes traffic signals and vehicle platooning at intersections through mixed integer linear programming, resulting in improved performance in terms of travel delay, throughput, fuel consumption, and emission compared to other approaches, which is verified by simulation \cite{liu2022single}.


\paragraph{Unsignalized intersections}
In unsignalized or uncontrolled intersections, collaborative control technology can help to improve safety and efficiency by enabling vehicles to communicate with each other and coordinate their movements \cite{xu2019cooperative}. This can help to reduce the likelihood of accidents caused by miscommunication or failure to yield, and enable smoother and more efficient traffic flow through the intersection \cite{pei2021optimal}.

For example, in order to coordinate the motion of vehicles at unsignalized intersections, Deng et al. propose a conflict duration graph-based coordination framework to resolve collisions and improve traffic capacity according to heuristic rules at signal-free intersections \cite{deng2020conflict}. 
Deng et al. further extend the research on collaborative control by proposing a bi-Level optimization method to coordinate the merging of CAVs at unsignalized intersections, aiming to achieve an optimal traffic schedule with dynamically-feasible and energy-efficient trajectories that improve space utilization and prevent spillbacks \cite{deng2023cooperative}.
In addition, Wang et al. present a digital twin framework for a collaborative control system at non-signalized intersections, which incorporates an enhanced FIFO slot reservation algorithm, a consensus motion control algorithm, a model-based motion estimation algorithm, and an augmented reality HMI \cite{wang2021digital}. 

Overall, this section discusses collaborative control techniques for CAVs, which are crucial for realizing their potential to improve transportation systems \cite{li2018humanlike, ohn2016looking, sun2021survey}. 
It covers enabling technologies, critical components, collaborative control methodologies, and potential applications. 
The enabling technologies provide the communication infrastructure for CAVs to share information and cooperate. Critical components such as longitudinal dynamics models, information flow topology, communication quality, and stability are necessary for collaborative control.
Collaborative control methodologies include centralized, decentralized, or mixed control mode, and utilize game-theoretic, optimization-based, or reinforcement learning-based methods. 
The potential applications of collaborative control include platooning, lane changing, intersection crossing, and more, which can enhance traffic efficiency, reduce emissions, and improve safety in one-, two-, or three-dimensional collaboration scenarios.

\section{Conclusion and Future Works}\label{sec:conclusion}




In this survey, we have provided a comprehensive review of vehicle control for autonomous vehicles and connected and automated vehicles. We began with vehicle state estimation, in particular, vehicle sideslip angle estimation, from the perspective of diverse sensor configurations and model features. 
We then discussed the trajectory tracking control for AVs with three types of autonomous driving control algorithms: feedback control without prediction, model predictive control with prediction, and learning-based control. 
We also analyzed different trajectory tracking algorithms from four main perspectives: model complexity, computation cost, optimal performance, and application scenarios.
We also explored the enabling techniques, critical components, collaborative control methodologies, and potential applications of cooperative driving control methods, which have the potential to significantly improve the efficiency, safety, and sustainability of the future transportation system.
In addition, we identify that future research on vehicle control from automated driving and cooperative driving automation will likely expand to the following avenues. 

1) \textbf{Multi-modality robust state estimation}: In addition to GNSS, LiDAR, and cameras can also be utilized to assist in vehicle state estimation to further improve the estimation accuracy. However, these sensors are susceptible to environmental factors, such as buildings, illumination, and weather. Furthermore, their sampling latency and frequency vary significantly. As a result, it is essential to investigate the confidence level of sensor signals in real-time and develop robust fusion algorithms.

2) \textbf{Robust trajectory tracking control with state uncertainty}: Thanks to the availability of multi-modal information, it is possible to estimate real-time three-dimensional vehicle attitude information. Consequently, the vehicle control model can be expanded from a two-dimensional plane to a three-dimensional space. Since the state estimation results may have a certain degree of uncertainty, it is crucial to account for the impact of state estimation uncertainty on the algorithm's performance while designing trajectory control algorithms.

3) \textbf{Addressing the control challenge of heterogeneous dynamics in mixed-autonomy vehicle platoons}:
The interaction and coordination between human-driven and AVs present a significant challenge to collaborative control techniques when encountering mixed-autonomy traffic. One of the biggest challenges in this area is the heterogeneous dynamics introduced by human-driven behavior in the vehicle platoon. 
To overcome this challenge, researchers and engineers will need to develop innovative control techniques that can adapt to the variability of human drivers while still maintaining safety and efficiency on the road. These techniques could include machine learning algorithms to predict human driver behavior, as well as hybrid control architectures that combine centralized and decentralized control techniques. This approach would allow AVs to adapt to the behavior of human-driven vehicles while still maintaining a level of coordination within the platoon.

4) \textbf{Resilience control techniques for CAVs in the presence of cyber-attacks and communication failures}: 
In a realistic and unpredictable driving environment, communication and sensing failures can occur, and adversarial entities may disrupt the system's operation. Therefore, it is necessary to integrate resilience control techniques into the collaborative control framework, enabling the system to continue to function and adapt to changing conditions even in the presence of faults, errors, or attacks.
Leveraging techniques from the field of cyber-physical systems will be an effective approach to designing resilient collaborative control techniques for CAVs. Additionally, considering security and privacy requirements will be critical. Designing communication, sensing, and control protocols with security and privacy in mind, such as utilizing encryption, authentication, and access control mechanisms, will be essential for ensuring the safety and performance of CAVs in the future.

5) \textbf{Developing a unified simulation platform and intelligent testing environments for safety evaluation of collaborative control techniques}:
The development of a unified simulation platform and intelligent testing environments is a critical step towards ensuring the safety of collaborative control techniques for CAVs. The platform will allow researchers and engineers to evaluate the effectiveness and safety of CAV control strategies in a virtual environment before they are implemented in the real world. To achieve this, a comprehensive understanding of CAV control strategies and their underlying technologies is necessary. The simulation platform should accurately model the behavior of CAVs and their interactions with other vehicles and infrastructure, while also employing machine learning techniques to enable accelerated testing and training with other safety-critical autonomous systems.	


\bibliography{Refs}
\bibliographystyle{ieeetr}

\end{document}